\documentclass[10pt,twocolumn,letterpaper]{article}

\usepackage[pagenumbers]{cvpr} 

\usepackage[toc,page]{appendix}

\usepackage{graphicx}
\usepackage{amsmath}
\usepackage{amssymb}
\usepackage{booktabs}
\usepackage{multirow}
\usepackage{color,soul}
\usepackage[ruled,vlined]{algorithm2e}
\usepackage{algorithmic}
\usepackage[flushleft]{threeparttable}
 
\SetKwInput{KwInput}{Input}                
\SetKwInput{KwOutput}{Output}              

\usepackage[pagebackref,breaklinks,colorlinks]{hyperref}
\usepackage{xcolor}
\usepackage[symbol, para]{footmisc}

\usepackage[capitalize]{cleveref}
\crefname{section}{Sec.}{Secs.}
\Crefname{section}{Section}{Sections}
\Crefname{table}{Table}{Tables}
\crefname{table}{Tab.}{Tabs.}

\DeclareMathOperator*{\argmax}{\arg\max}
\DeclareMathOperator*{\argmin}{\arg\min}

\def\cvprPaperID{10096} 
\def\confName{CVPR}
\def\confYear{2022}

\newcommand{\specialcell}[2][c]{%
  \begin{tabular}[#1]{@{}c@{}}#2\end{tabular}}
\newcommand{\centered}[1]{\begin{tabular}{c} #1 \end{tabular}}

\begin{document}
\title{\vspace{-1.4cm}DTA: Physical Camouflage Attacks \\ using Differentiable Transformation Network}

\author{Naufal Suryanto $^{1, \ddag}$, Yongsu Kim $^{1,2, \ddag}$, Hyoeun Kang $^{1}$, Harashta Tatimma Larasati $^{1,4}$, Youngyeo Yun $^{1}$ \\
Thi-Thu-Huong Le $^{1,5}$, Hunmin Yang $^{3}$, Se-Yoon Oh $^{3}$, Howon Kim $^{1,2, \textasteriskcentered}$\\
{\small $^{1}$Pusan National University, South Korea; $^{2}$SmartM2M, South Korea; $^{3}$Agency for Defense Development (ADD), South Korea;} \\ 
{\small $^{4}$Institut Teknologi Bandung, Indonesia; $^{5}$Hung Yen University of Technology and Education, Vietnam}\\
{\small \url{https://islab-ai.github.io/dta-cvpr2022/}}
}


\maketitle
\begin{abstract}
    To perform adversarial attacks in the physical world, many studies have proposed adversarial camouflage, a method to hide a target object by applying camouflage patterns on 3D object surfaces.
    For obtaining optimal physical adversarial camouflage, previous studies have utilized the so-called neural renderer, as it supports differentiability. However, existing neural renderers cannot fully represent various real-world transformations due to a lack of control of scene parameters compared to the legacy photo-realistic renderers. 
    In this paper, we propose the Differentiable Transformation Attack (DTA), a framework for generating a robust physical adversarial pattern on a target object to camouflage it against object detection models with a wide range of transformations. 
    It utilizes our novel Differentiable Transformation Network (DTN), which learns the expected transformation of a rendered object when the texture is changed while preserving the original properties of the target object. 
    Using our attack framework, an adversary can gain both the advantages of the legacy photo-realistic renderers including various physical-world transformations and the benefit of white-box access by offering differentiability. 
    Our experiments show that our camouflaged 3D vehicles can successfully evade state-of-the-art object detection models in the photo-realistic environment (i.e., CARLA on Unreal Engine). Furthermore, our demonstration on a scaled Tesla Model 3 proves the applicability and transferability of our method to the real world. \footnotetext[3]{Equal contribution} \footnotetext[1]{Corresponding author}
    
\end{abstract}
\begin{figure}
  \begin{subfigure}{0.49\columnwidth}
    \centering
    \includegraphics[width=\columnwidth]{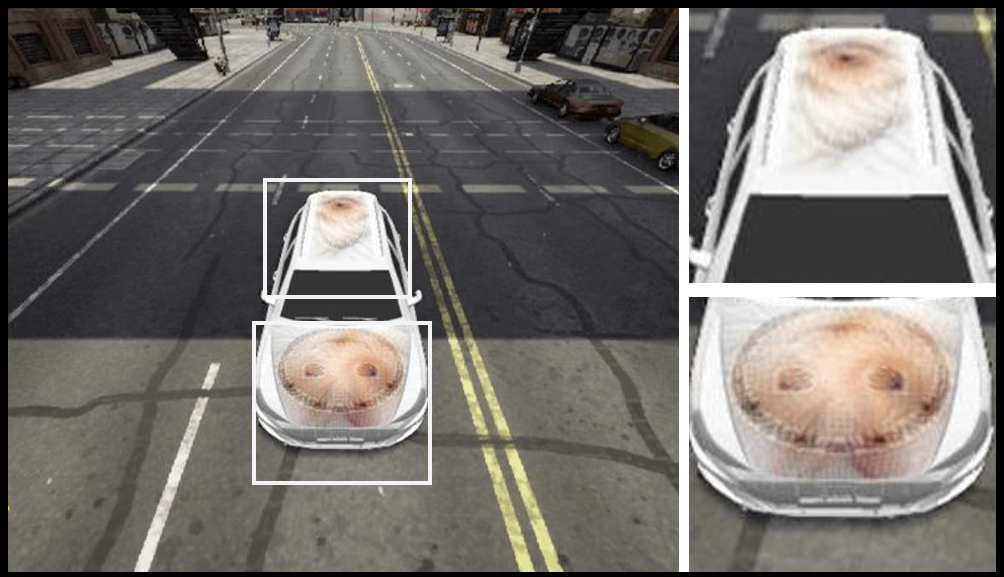}
    \caption{} \label{fig:das_weakness}
  \end{subfigure}\hfill
    \begin{subfigure}{0.49\columnwidth}
    \centering
    \includegraphics[width=\columnwidth]{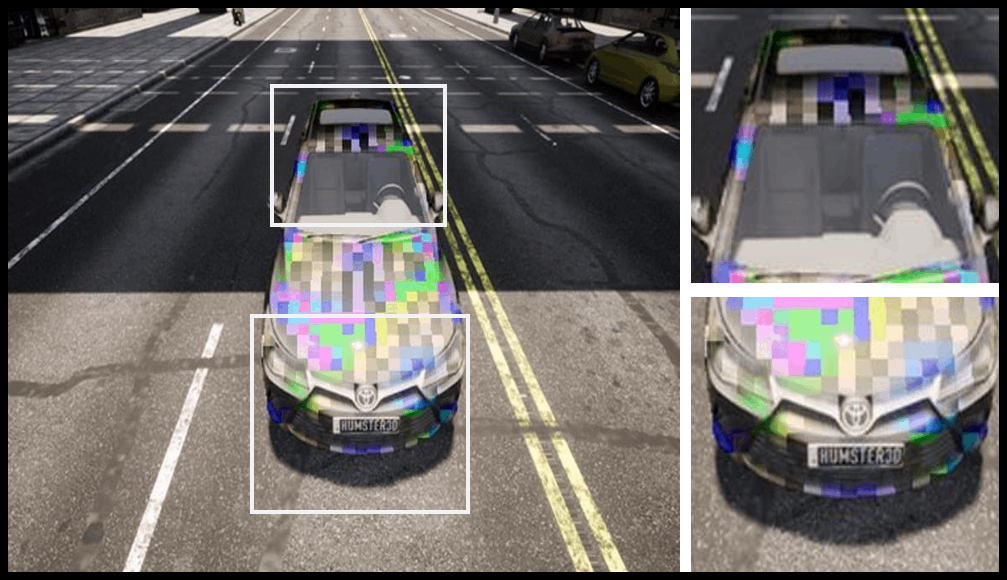}
    \caption{}
  \end{subfigure}
  \begin{subfigure}{\columnwidth}
    \centering
    \includegraphics[width=\columnwidth]{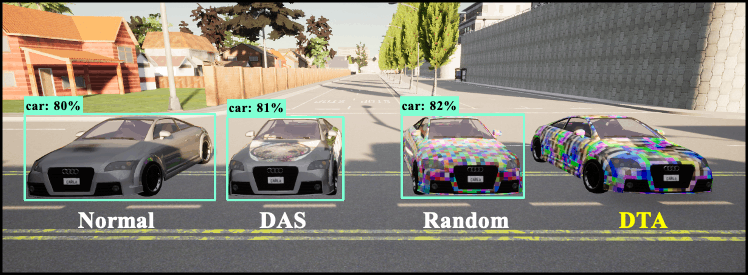}
    \caption{}
  \end{subfigure}
 
  \caption{
    \textbf{(a) Neural renderer used in prior work (Dual Attention Suppression (DAS)) \cite{dasattack}.}
    Although it is differentiable, it possesses a limitation in physical properties representation (e.g., transparency for windshield) and lacks background blending (e.g., shadowing) because the object and the background scene are rendered separately. 
    \textbf{(b) Our Differential Transformation Network (DTN).}
    DTN considers both differentiability and photo-realistic aspects by learning the correct transformation when the object's texture is changed. As shown, transparency for the windshield and shadowing at the bottom and the top of the car are rendered correctly.
    \textbf{(c) Comparison of detection results using different textures \textit{(normal, DAS, random, DTA)} on photo-realistic environment.}
    Unlike other examples, the adversarial camouflage generated by our DTA framework successfully evades detection.
    } 
  \label{fig:eotnet_contribution}
\end{figure}

\section{Introduction} \label{sec:intro}
Deep neural networks (DNNs), despite their renowned capability for solving computer vision tasks \cite{classification, mobilenets, recognition}, have been proven vulnerable to adversarial examples \cite{Szegedy2014IntriguingPO}. That is, carefully crafted inputs may cause DNN models to misrepresent a seemingly obvious image to the human eye, giving incorrect prediction results. A deliberate act by an adversary to take advantage of this weakness, namely the adversarial attack, has captured the attention of many in the past few years. Its potential applicability in not only the digital domain but also the physical domain has drawn significant interest.

Compared to digital attacks, physical adversarial attacks are more difficult to launch since they must account for various physical constraints and conditions (e.g., lighting, camera pose, and occlusion). However, the fully physical attack experiments in the real world such as \cite{EoT, upcattack, ShapeShifter, Fooling} are extremely time consuming and expensive. Therefore, various studies have been conducted through simulation of the physical world in the digital environment by using legacy photo-realistic rendering software, such as Unreal Engine \cite{unrealengine} and AirSim \cite{airsim}, facilitating parameter control. Examples of methods to craft physical adversarial camouflage, i.e., adversarial attack variant that focuses on hiding an object by fully covering the target object, in simulators can be found in \cite{CAMOU, ER}. However, since such simulators are non-differentiable, the attacks employ a black-box approach such as utilizing a clone network \cite{CAMOU} or a genetic algorithm \cite{ER}, yielding an inevitably lower attack performance than the white-box counterpart.

To obtain the advantage of differentiability, more recent methods \cite{dasattack, dpaattack, fcaattack} have proposed the use of neural renderers for generating adversarial camouflage. However, the existing neural renderers (e.g., \cite{NeuralMeshRenderer0}) can only support the generation of foreground objects; hence, background images are still handled by the legacy photo-realistic renderers. 
As a result, they simply attach the generated target object to the background image, yielding inaccurate foreground-background blending effects, such as shadow casting and light reflection, as shown in Fig. \ref{fig:das_weakness}. Although workaround efforts, such as masking for handling occlusion \cite{fcaattack}, have been proposed, the overall adversarial camouflage results using existing neural renderers are still inferior in terms of photo-realistic attributes.

Motivated by the challenge faced in prior works, we develop an attack framework 
that takes advantage of the differentiability in neural renderers without compromising the photo-realistic properties of the target object. In particular, the framework leverages our novel neural rendering technique, which learns the representation of various scene properties (e.g., object material, lighting effects, and shadows) from legacy photo-realistic renderers. As a result, truly robust physical adversarial camouflage, verified from our experiment using both a photo-realistic simulation and a real-world example, can be obtained.

Our contributions can be summarized as follows: 
\begin{itemize}
\item We present the Differentiable Transformation Attack (DTA), a framework for generating robust physical adversarial camouflage on 3D objects. It combines the advantages of a photo-realistic rendering engine with the differentiability of our novel rendering technique.

\item We propose the Differentiable Transformation Network (DTN), a brand-new neural renderer that learns the transformations of an object when the texture is changed while preserving its original parameters for a realistic output that resembles the original.
\item Our DTN can be embedded as an extension to provide differentiability to any rendering software (e.g., Unreal Engine \cite{unrealengine}), enabling the use of any gradient-based method. 
\item We demonstrate that the adversarial camouflage generated from DTA is robust and applicable for evading pre-trained object detection models under various transformations in both simulations and the real world.
\item Our attack method, DTA, outperforms previous works in terms of our evaluation of target object detection models and transferability to other models.  
\end{itemize}

\section{Related Works}

\textbf{Physical Adversarial Attack} To launch an adversarial attack in the physical world, one of the most notable proposals is the Expectation Over Transformation (EOT) \cite{EoT}, which generates robust adversarial examples under various transformations, such as viewing distance, angle, and lighting condition. Most of the recent physical adversarial attack methods employ EOT-based algorithms to make the attack performance robust in the real world.

\textbf{Adversarial Camouflage} Physical perturbation \cite{ShapeShifter} and patch-based methods \cite{DPatch, Fooling, PhysicalPatch} targeting planar and rigid objects were mainly proposed for real-world adversarial camouflage attacks. Subsequently, Huang et al. \cite{upcattack} proposed Universal Physical Camouflage (UPC) to generate adversarial camouflage that also covers non-planar and non-rigid objects. However, these methods can only be applied to a segment of an object and, due to their nature, can only attack at certain viewing angles.

A more recent approach of adversarial camouflage involves manipulating the color texture pattern of the target 3D object to degrade the detection performance of object detectors. This technique has the advantage of the ability of attack from any viewing angle by covering all parts of the object. Initially, black-box attack methods were commonly proposed since the rendering process, including texture mapping, is non-differentiable. For instance, Zhang et al. \cite{CAMOU} proposed CAMOU, an adversarial camouflage method to hide vehicles from detectors by training a clone network that imitates both applying camouflage to vehicles and detecting the camouflaged vehicles. Meanwhile, Wu et al. \cite{ER} presented adversarial camouflage based on genetic algorithms to be applied on vehicle surfaces so that it is not recognizable by detectors in the CARLA simulator \cite{carla}. 

\begin{figure*}
\begin{center}
\centerline{\includegraphics[width=0.85\textwidth]{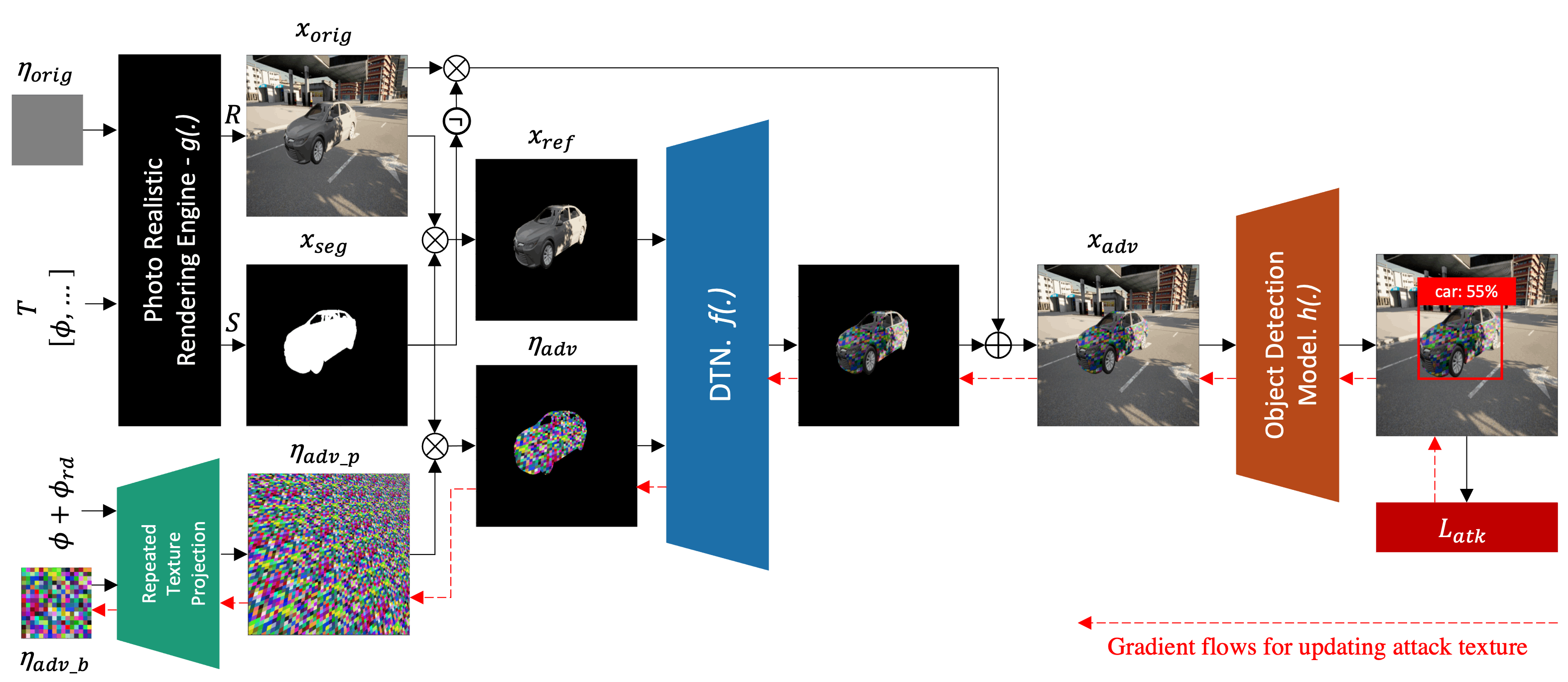}}
\caption{DTA framework for generating robust adversarial texture.}
\label{fig:EOTnet_attack_pipeline}
\end{center}
\end{figure*}

\textbf{Neural Renderer-based Methods} To gain white-box access \textemdash and in turn, improve the attack performance, there is a rising trend of leveraging the differentiability intrinsic in neural renderers for the adversarial camouflage generation, such as in \cite{dasattack, fcaattack, dpaattack}.
For example, \cite{dasattack} proposed the Dual Attention Suppression (DAS) attack to generate natural adversarial camouflage using a Neural 3D Mesh Renderer \cite{NeuralMeshRenderer0} by suppressing the model and human attention. However, existing neural renderers pose limitations on handling complex 3D interactions among scene properties. Inevitably, the resulting camouflage may not properly address the photo-realistic effects of the physical world.

To tackle these issues, we design an attack framework that combines the best of both worlds; that is, it has the differentiability of a neural renderer while retaining the photo-realistic attributes of the object. Our attack, namely the  Differentiable Transformation Attack (DTA), utilizes our novel neural rendering model called Differentiable Transformation Network (DTN), which learns the representation of various scene properties from a legacy photo-realistic renderer, giving a more applicable solution in the physical world.

\begin{figure*}
\begin{center}
\centerline{\includegraphics[width=0.85\textwidth]{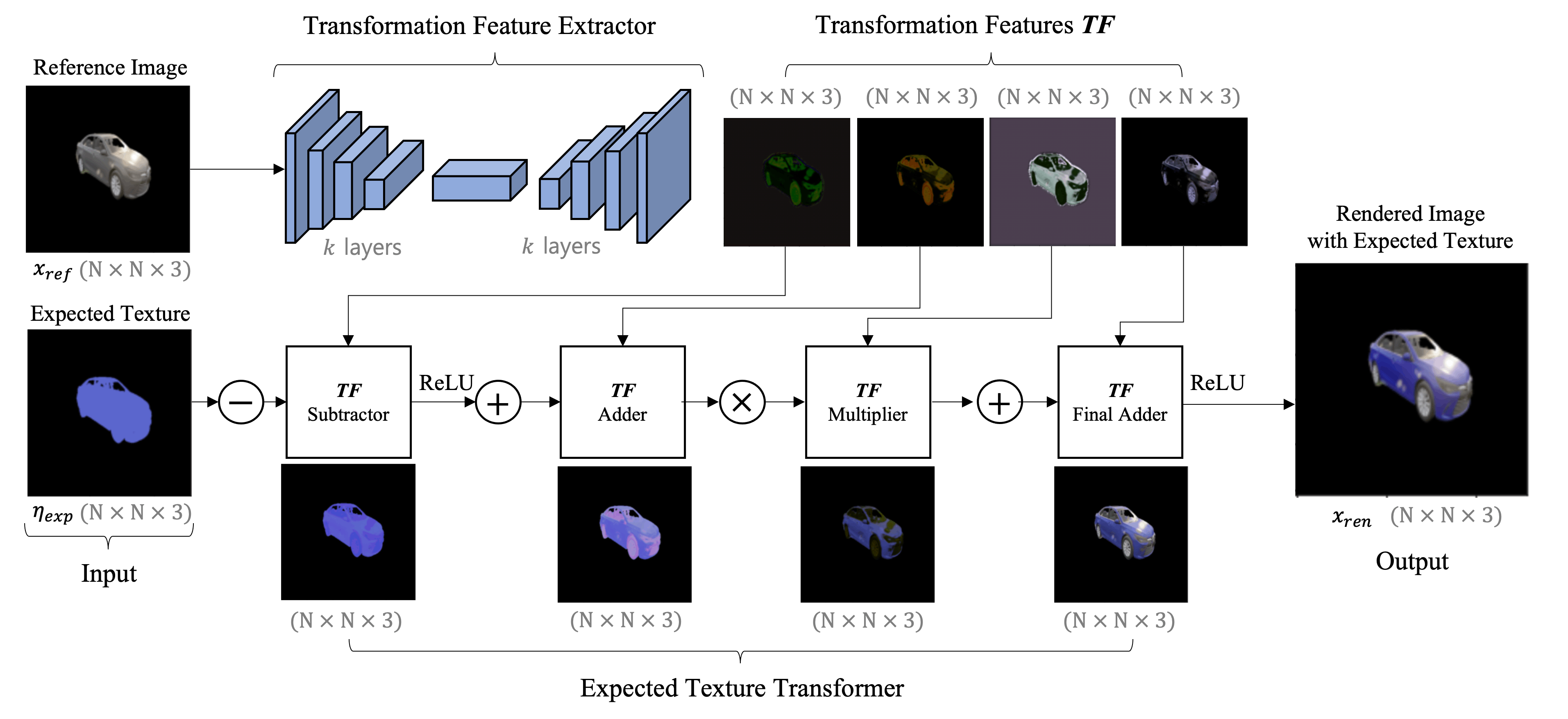}}
\caption{DTN architecture, network for learning expected transformation.}
\label{fig:DTN_architecture}
\end{center}
\end{figure*}

\section{Methodology}
In this section, we first describe DTN, our differentiable renderer that learns the expected transformation of projecting a specific pattern from a photo-realistic rendering engine. Furthermore, we describe DTA, the attack framework utilizing DTN for generating robust adversarial camouflage. 

\subsection{Problem Definition}
The final goal of our proposed method is to generate a robust adversarial pattern to be applied to 3D objects in the simulated physical world to significantly degrade the detection score of that object in a variety of transformations, such as object materials, camera poses, lighting conditions, and background interactions.

Let $h_\theta$ be a hypothesis function for the object detection task, that satisfies $h_\theta(x)=y$. The notion $x$ as the input denotes the 2D image, which includes the target object generated from the rendering process, and $y$ as the output denotes the label of the detection result of the corresponding target object. The goal of our proposed method is to generate adversarial example $x_{adv}$, which satisfies $h_\theta(x_{adv})\neq y$, by modifying the texture pattern of the target object. Suppose $L(h_\theta(x), y)$ is a loss function applied to $h_\theta$ that enables detection of an object in $x$ as $y$. We can generate $x_{adv}$ by solving Eq. \ref{eq:loss_x_adv}.
\begin{equation}
    \argmax_{x_{adv}}L(h_\theta(x_{adv}), y)
    \label{eq:loss_x_adv}
\end{equation}
Unlike 2D adversarial examples where we can directly modify the input image pixels, applying texture in 3D objects requires a rendering process with many parameters affecting the final image, such as shadows and light reflection.

Suppose $R$ is a rendering function used in the photo-realistic rendering engine $g$, that satisfies Eq. \ref{eq:rendering_eq_fix},
\begin{equation}
    R(T, \eta) = x
    \label{eq:rendering_eq_fix}
\end{equation}
where $T$ is a transformation matrix encoding various transformations, such as camera pose $\phi$, lighting conditions, meshes, and material properties, as well as the target object and its location, whereas $\eta$ is a texture that will be applied to the target object.
If $R$ is differentiable, we can find a robust adversarial texture $\eta_{adv}$ that works in a wide variety of $T$ using EOT \cite{EoT} method. However, since $R$ is not always differentiable, we propose a neural network $f_\omega$ that learns the texture transformations by solving Eq. \ref{eq:EOTnet_eq_fix},
\begin{equation}
    f_\omega(x_{ref}, \eta_{exp}) = x_{ren}
    \label{eq:EOTnet_eq_fix}
\end{equation}
where $x_{ref}$ is the reference image obtained from the rendering function containing the transformation information, $\eta_{exp}$ is the expected texture variable, and $x_{ren}$ is the rendered image with the expected texture. When $\eta_{exp}$ is the same texture used in $x_{ref}$, then $x_{ref}=x_{ren}$. Since $f$ is differentiable, we can generate adversarial texture $\eta_{adv}$, which satisfies $f_\omega(x_{ref}, \eta_{adv})=x_{adv}$, by solving Eq. \ref{eq:loss_e_adv}.
\begin{equation}
    \argmax_{\eta_{adv}}L(h_\theta(f_\omega(x_{ref}, \eta_{adv})), y)
    \label{eq:loss_e_adv}
\end{equation}

\subsection{DTA Framework}
The proposed DTA framework uses DTN, a neural network designed to solve the problem as described in Eq. \ref{eq:EOTnet_eq_fix}.
DTN learns the transformation of the rendered object in $x_{ren}$ given a reference image $x_{ref}$ and expected texture $\eta_{exp}$.
It relies on the photo-realistic image synthesized from a non-differentiable renderer to produce a differentiable version of the photo-realistic reference image after applying the expected texture.
Furthermore, the differentiability of DTN is used to generate the robust adversarial pattern applied to the target object as the rendered adversarial texture for camouflaging the object detector. As a result, DTN provides a white-box ability for lowering specific attack loss.
As depicted in Fig. \ref{fig:EOTnet_attack_pipeline}, the DTA framework comprises four components: a photo-realistic rendering engine, a Repeated Texture Projection function, DTN, and the target object detection model.

\textbf{Photo-Realistic Rendering Engine}
In our proposed DTA framework, the photo-realistic rendering engine is any software that can produce a photo-realistic image that is similar to the real physical world.
One example is a game engine. It can be used for building a fully simulated physical world and synthesizing a photo-realistic image \textemdash even some can also synthesize semantic segmentation image, thanks to its detailed features and interactivity. In comparison, the interactivity has not been properly addressed in the existing differentiable renderers. 
Then, DTN is embedded as an extension to enable differentiability of the texture space, allowing adversarial texture generation for the target object.

\textbf{DTN}
This technique uses photo-realistic RGB images synthesized from a rendering engine as the input reference image $x_{ref}$. The reference image only contains the masked target object where the expected texture $\eta_{exp}$ will be applied.
This masked image enables DTN to solely focus on learning and applying the transformation of the target object while ignoring the background.

In terms of architecture, DTN mainly consists of the Transformation Feature Extractor and Expected Texture Transformer, as illustrated in Fig. \ref{fig:DTN_architecture}. The Transformation Feature Extractor is a convolutional autoencoder-like neural network that learns to extract transformation features of reference image $x_{ref}$ and encode them as stacked transformation features $TF$. They are then sliced and used for transforming expected texture $\eta_{exp}$ into a rendered image $x_{ren}$.

The final output of the Transformation Feature Extractor is $TF$. It has a $N \times N \times 12$ shape, where $N$ is the resolution of input and output images and $12$ is the last channel representing the four stacked RGB transformation features used to transform the expected texture to the image rendered by the Expected Texture Transformer. 
$TF$ will have the same value no matter what the expected texture is. The idea is to prevent the Transformation Feature Extractor from overfitting because the expected texture is designed to never be shown directly as input to the network.  

The transformation of $\eta_{exp}$ into $x_{ren}$ is performed using basic mathematical operations, such as subtraction, addition, and multiplication by each slice of $TF$. The design of this expected texture transformer is assumed to cover all kinds of basic transformations.

\begin{figure}
\centerline{\includegraphics[width=\columnwidth]{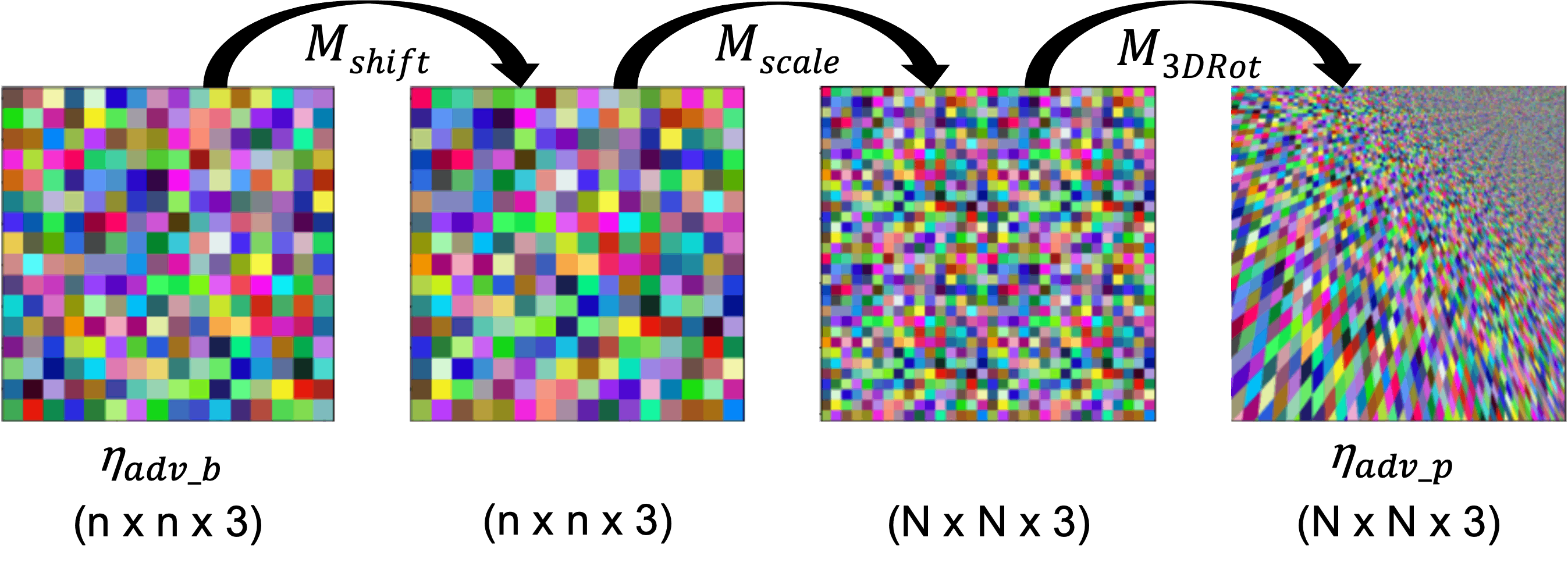}}
\caption{Sequence of transformation in Repeated Texture Projection Function.}
\label{fig:repeated_texture_projection}
\end{figure}

\textbf{Repeated Texture Projection Function}
We propose a repeated pattern as our final attack camouflage texture. The repeated pattern has several advantages, such as ease of application because the texture can be used to cover the object while ignoring the texture mapping. 
The repeated pattern will also look more similar when we vary the viewing angle, creating a more robust attack.

In Figs. \ref{fig:EOTnet_attack_pipeline} and \ref{fig:repeated_texture_projection}, we define an $n \times n$ adversarial pattern $\eta_{adv\_b}$ that will be transformed into $\eta_{adv\_p}$ until it has the same size as $N \times N$ DTN input requirement. We propose a Repeated Texture Projection function for simply projecting the repeated pattern based on the same camera pose $\phi$ used by the photo-realistic rendering engine. We add random pose $\phi_{rd}$ for adding a variety of transformations when it is used to generate the adversarial pattern, which is expected to increase the attack texture robustness and anticipate projection error. The Repeated Texture Projection function contains a sequence of operations for transforming the adversarial pattern with transformation matrix $M$.

The transformation matrix $M$ used by repeated texture projection function covers shift, scale, and 3D rotation on 2D image operations.
We can write it as Eq. \ref{eq:texture_projection_matrix},
\begin{equation}
   \eta_{adv\_p} = M_{3DRot} \cdot M_{scale} \cdot M_{shift} \cdot \eta_{adv\_b}
    \label{eq:texture_projection_matrix}
\end{equation}
where $M_{shift}$ is a shift operation that determines the pattern order or initial location, $M_{scale}$ is a scale operation that determines how large the texture will be when it is resized, and $M_{3DRot}$ is a 3D rotation operation that determines how the 2D texture is rotated along the 3D axis. Each matrix $M$ is calculated or calibrated based on a given camera pose $\phi + \phi_{rd}$ such that the projection covers the majority of the target's flat surface. Fig. \ref{fig:repeated_texture_projection} describes the result of each transformation sequence. For filling points outside boundaries, we use wrap mode, which extends the output by wrapping around the opposite edge, giving a repeated texture effect.

\subsection{Framework Procedure}

\textbf{DTN Model Training}
Before using DTA for generating the adversarial pattern, DTN is trained with the dataset generated by the photo-realistic rendering engine. We set the dataset with two inputs, $x_{ref}$ and $\eta_{exp}$, and one output, $x_{ren}$. First, we select a set of random flat color texture and predefined transformations. Then, we use the rendering engine to produce the photo-realistic images that will later be used as reference image $x_{ref}$, expected texture $\eta_{exp}$, and ground truth of rendered image $x_{ren}$.
We use flat color texture as the expected texture so that there is no error caused by texture mapping during the training.

For each training example, $x_{ref}$ only contains the masked target object applied with specific color and transformations. $\eta_{exp}$ contains the flat color texture that will be applied to the target object and has a masked shape of the target object. We set the ground-truth of $x_{ren}$ as the result of applying the flat color texture used in $\eta_{exp}$ to the target object using the photo-realistic rendering engine. In this case, the same transformations used in $x_{ref}$ except the texture are applied to the target object. Additionally, we utilize binary cross-entropy loss as per-pixel construction loss for training the DTN. The detailed algorithm of the DTN training process is described in the supplementary material.

\textbf{DTA Attack Phase}
In the attack phase, the attack goal is to minimize the original target confidence score, which prevents the object detector from detecting the target object correctly.
Since the object detection model outputs multiple boxes and class confidence scores, we just take the maximum target object confidence scores $C$ and measure the log loss when we set the ground truth to zero.
We can write the attack loss $L_{atk}$ representing the above process as Eq. \ref{eq:EOTnet_attack_loss}.
\begin{equation}
   L_{atk}(h(x)) = \mathbb{E}_{t\mathtt{\sim}T} [-log(1 - max(C(h(x))))]
    \label{eq:EOTnet_attack_loss}
\end{equation}
\begin{equation}
    \eta_{adv} = \argmin_{\eta}L_{atk}(h(f_w(x_{ref}, \eta)))
    \label{eq:DTN_atk_texture}
\end{equation}
Minimizing $L_{atk}$ has the same effect as solving Eq. \ref{eq:loss_e_adv}. We can use the differentiability of the full attack pipeline to find the best adversarial pattern $\eta_{adv}$ that minimizes the attack loss by updating the $\eta_{adv}$ based on the loss gradient. Eq. \ref{eq:DTN_atk_texture} describes the calculation of the best adversarial pattern $\eta_{adv}$. 
The full pipeline of the DTA framework for generating a robust attack pattern can be seen in Fig. \ref{fig:EOTnet_attack_pipeline} and Algorithm \ref{alg:attack_dta}. In Algorithm \ref{alg:attack_dta}, DTN $f_w$ is the model that has completed the training process.

\begin{algorithm}[]
    \caption{Generating attack texture using DTA}
    \label{alg:attack_dta}
\begin{algorithmic}
    \STATE {\bfseries Input:} Transformation set $T = \{t^{(1)},\dots,t^{(M)}\}$, Base flat color texture $C$, Rendering function $R$, Segmentation function $S$, Repeated texture projection function $P$, DTN $f_w$
    \STATE {\bfseries Output:} Attack texture $\eta_{adv\_b}$ \\ 
    (1) Export $X_{orig}$, $X_{seg}$, and $X_{ref}$ from the photo-realistic rendering engine
    \FOR {$m = 1$ to $M$}
    \STATE $X_{orig}^{(m)} \gets R(t^{(m)}, C)$
    \STATE $X_{seg}^{(m)} \gets S(t^{(m)})$
    \STATE $X_{ref}^{(m)} \gets X_{orig}^{(m)} \times X_{seg}^{(m)}$
    \ENDFOR \\
    (2) Generate attack texture
    \STATE Initialize $\eta_{adv\_b}$ with random values
    \FOR {number of training iterations}
    \STATE Sample minibatch of each $b$ samples $x_{ref} \in X_{ref}$, $x_{seg} \in X_{seg}$
    \STATE Derive $\phi$ corresponding to each $x_{ref}$ from $T$
    \STATE $\eta_{adv\_p} \gets P(\eta_{adv\_b}, \phi + \phi_{rd})$
    \STATE $\eta_{adv} \gets \eta_{adv\_p} \times x_{seg}$
    \STATE Derive $x_{adv}$ from $f_w(x_{ref}, \eta_{adv})$
    \STATE $x_{adv} = x_{adv} + (x_{orig} \times \neg x_{seg})$
    \STATE Calculate $L_{atk}(h(x_{adv}))$ by Eq. \ref{eq:EOTnet_attack_loss}
    \STATE Update $\eta_{adv\_b}$ for minimizing $L_{atk}(h(x_{adv}))$ via backpropagation
    \ENDFOR
\end{algorithmic}
\end{algorithm}

\section{Experiments}

\subsection{Implementation Details} 

\textbf{DTA Framework} We utilize TensorFlow 2 \cite{tf} for implementing our DTA framework, except for the photo-realistic renderer, in which we use CARLA \cite{carla} simulator on Unreal Engine 4 \cite{unrealengine}. CARLA provides ready-to-use APIs and digital assets (e.g., urban layouts, buildings, and vehicles) to simulate the physical world required for self-driving car research experiments. In our case, we modify the original CARLA code to allow modification of car texture, which is required for dataset generation and texture evaluation. Additionally, we employ world-aligned texture in Unreal for repeated texture implementation instead of the original car's UV mapping.

\textbf{Target Object} We choose Toyota Camry as our target object for adversarial camouflage generation and evaluation, Audi TT for camouflage transferability evaluation, and Tesla Model 3 for real-world evaluation.

\textbf{Datasets} For DTN model training and evaluation, we select a map in CARLA and randomly choose 75 spawn locations for generating both training and validation datasets, and another 50 for generating testing datasets. We spawn the target car at each location and capture the images using cameras with 5m distance, 15-degree pitch, and every 45-degree rotation as a single transformation. For each transformation, we sequentially change the car texture with 50 random flat color textures. Our train and test datasets use different transformations and colors.
For adversarial pattern generation, we select 250 spawn locations in the same map and generate datasets of the target object with the flat color texture used as the reference image during DTN training.

\textbf{Evaluation Metrics} To measure DTN's accuracy in predicting the transformation (with respect to the ground truth from the photo-realistic renderer), we use the model loss (i.e., binary cross-entropy) and mean squared error (MSE). To evaluate the performance of our proposed DTA, we use Average Precision@0.5, a commonly used metric for evaluating object detection models, including in the previous works. Furthermore, since our framework's main objective is to lower the target object's confidence score, it is also considered as another evaluation metric.

\textbf{Target Models} For ease of reproducibility, we choose the state-of-the-art COCO pre-trained object detection model \cite{tfod} and employ both EfficientDetD0 \cite{efficientdet} and YOLOv4 \cite{yolov4} as the target models. Furthermore, we evaluate the transferability of the generated pattern to other models (i.e., SSD \cite{ssd}, Faster R-CNN \cite{fasterrcnn}, and Mask R-CNN \cite{maskrcnn}).

\textbf{Compared Methods} We compare our adversarial camouflage with previous works on 3D physical attacks: CAMOU \cite{CAMOU}, ER \cite{ER}, UPC \cite{upcattack}, DAS \cite{dasattack}, and an additional random pattern. 
However, UPC and DAS have different settings to recreate in our environment; thus, we only evaluate them on the transferability experiment. 
Details on the parameters and setup are available in the supplementary material.

\textbf{DTN Parameters} We employ a batch size of 32, 25 epochs, the Adam optimizer \cite{kingma2014adam}, and the same random seeds on each trial as the fixed parameters for training all networks. The input and output size of DTN are 512$\times$512$\times$3 to match the target object detection input size. 

\textbf{DTA Framework Parameters} We select the 16$\times$16 texture size as the adversarial camouflage following CAMOU's best implementation. We employ 32 and 16 batch sizes for generating adversarial camouflage on EfficientDetD0 and YOLOv4, respectively. Each generation uses 200 epochs. 

\subsection{DTA Experiments}

\begin{figure}
\centerline{\includegraphics[width=\columnwidth]{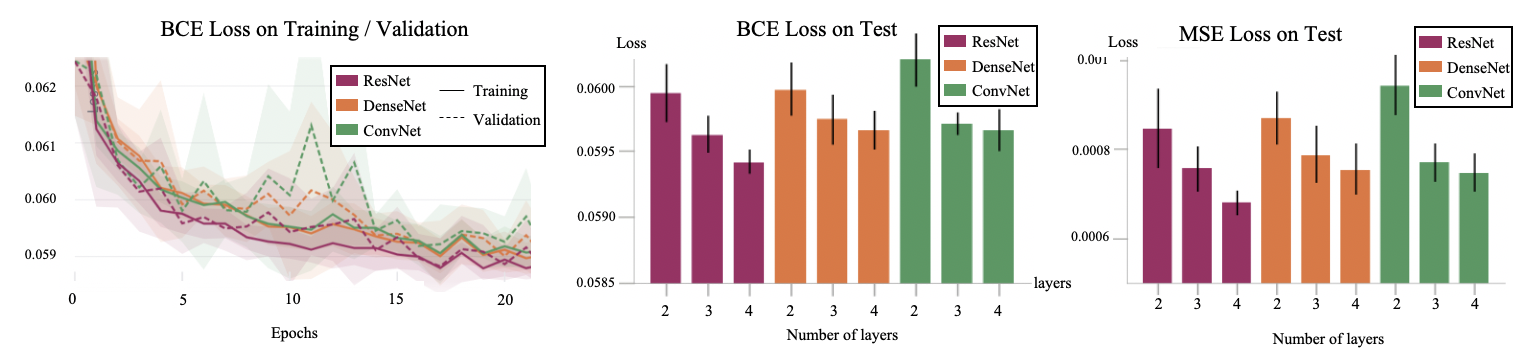}}
\caption{DTN model evaluations with different architectures [ResNet (red), DenseNet (orange), ConvNet (green)] and numbers of layers $k = [2, 3, 4]$.}
\label{fig:dtn_evaluation}
\end{figure}

\begin{figure}
\centerline{\includegraphics[width=\columnwidth]{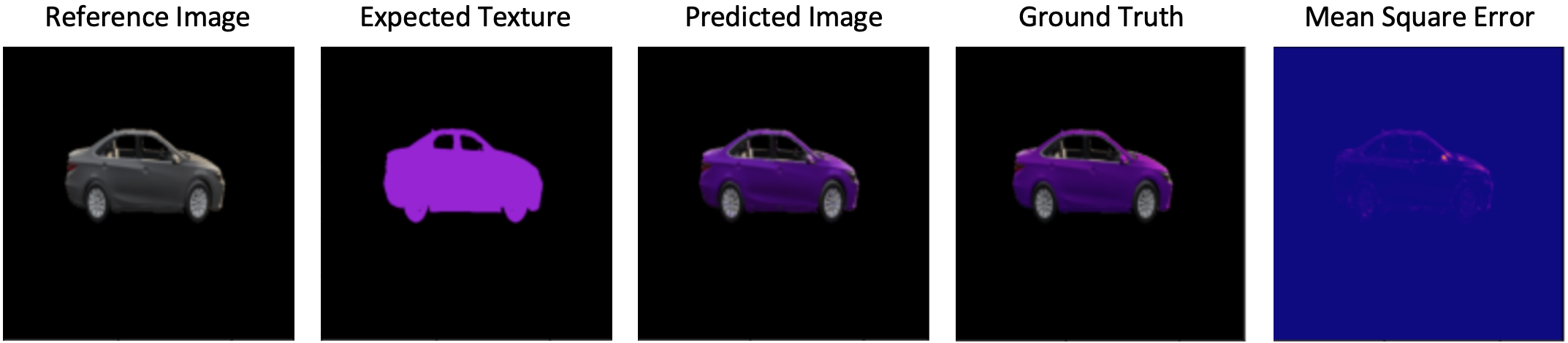}}
\caption{Example of DTN prediction results.}
\label{fig:dtn_result}
\end{figure}

\begin{figure}
\centerline{\includegraphics[width=\columnwidth]{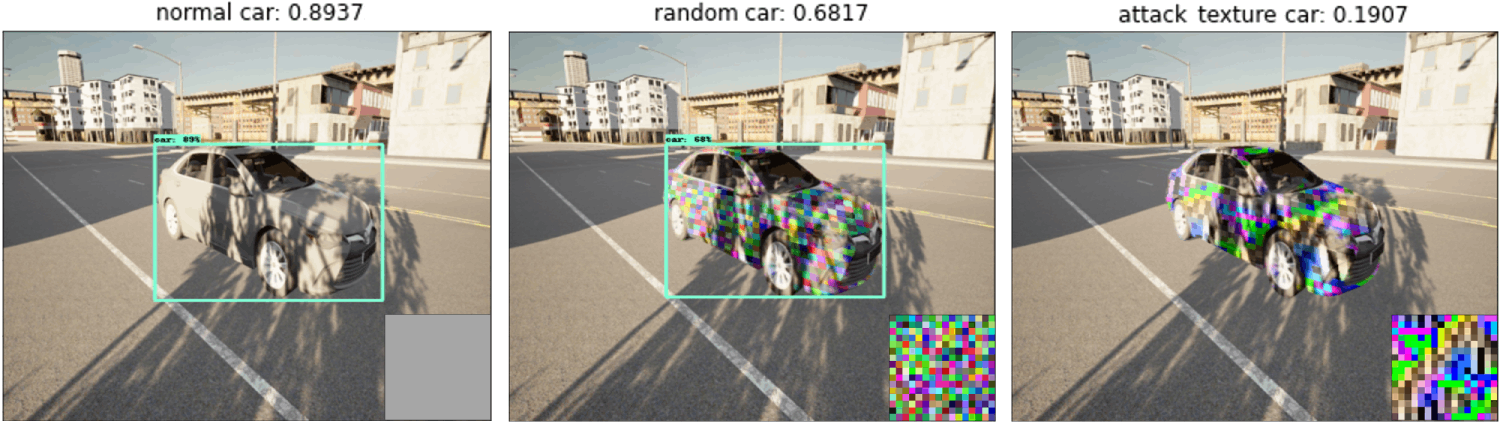}} 
\caption{DTN texture rendering results with detection. From left to right: normal texture (detected as car), random texture (detected as car), attack texture (mis-detection). The car's confidence score for attack textures is the lowest at 0.1907.}
\label{fig:adversarial_pattern_by_dtn}
\end{figure}

\textbf{DTN Evaluation} We employ $k = [2, 3, 4]$ layers for the Transformation Feature Extractor and utilize either residual \cite{resnet} or dense connection \cite{densenet} on the encoder besides a plain CNN. 
From Fig. \ref{fig:dtn_evaluation}, we can infer that DTN with residual connection has the overall lowest average test loss and MSE, followed by DTN with dense connection. 
In our experiments, increasing the number of layers to four lowers the overall model loss, with the best model achieving a 0.05942 average test model loss and 0.00068 average test MSE, resulting in an accurate rendering prediction as shown in Fig. \ref{fig:dtn_result}.
More detailed evaluations, graphs, and prediction examples are included in the supplementary material.

\textbf{Adversarial Camouflage Generation} During the adversarial camouflage generation, we use DTN with ResNet and $k$ of 4. We first generate a random pattern texture and optimize it to lower the attack loss using the full pipeline of the DTA framework. 
Fig. \ref{fig:adversarial_pattern_by_dtn} shows how the texture is rendered and detected by DTA. As shown, after the completion of attack camouflage generation, the target confidence score is much lower, which may cause miss-detection. Moreover, it can be seen that a sole random pattern texture is not sufficient for camouflaging the object detector.

\begin{figure}
\centerline{\includegraphics[width=\columnwidth]{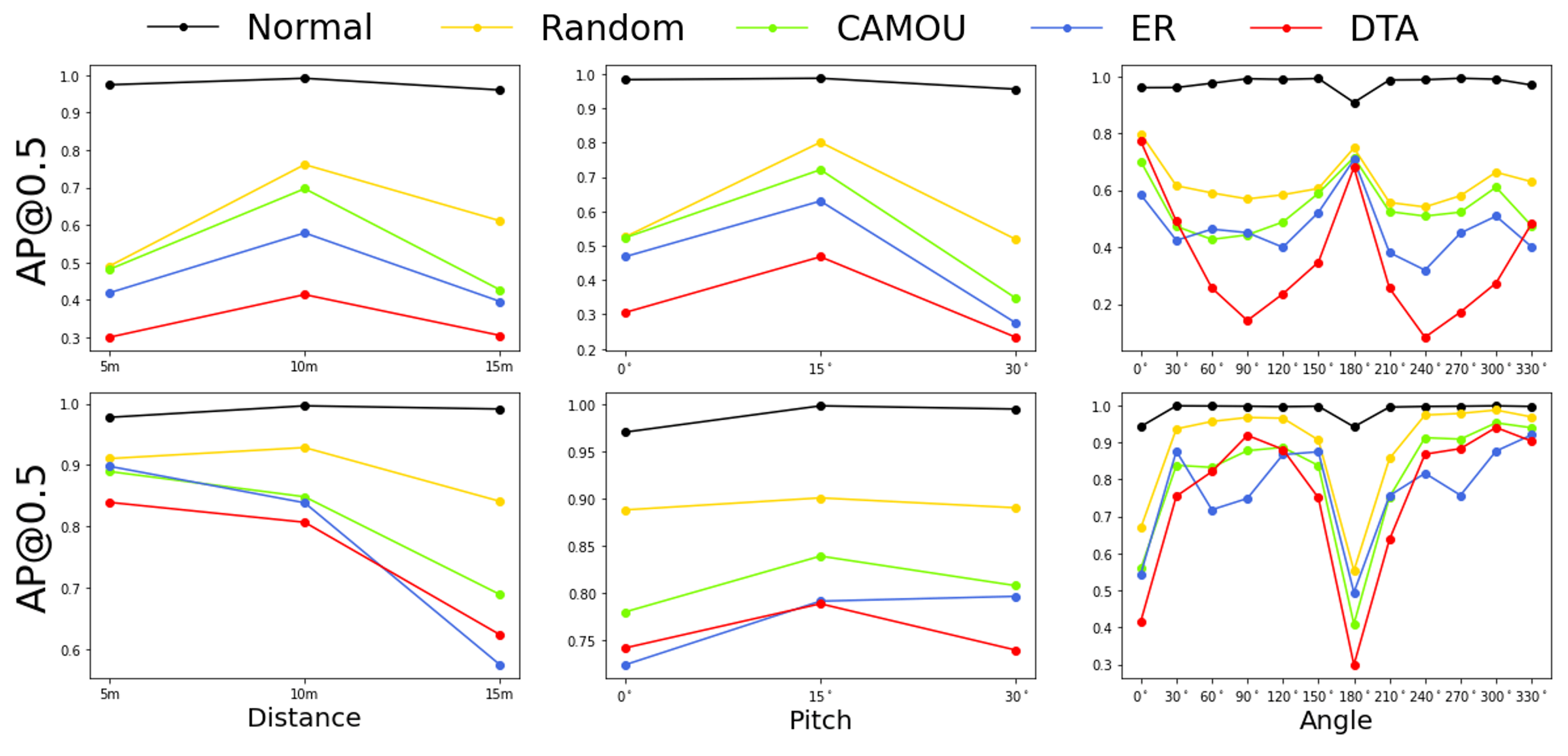}} 
\caption{Average Precision@0.5 of various rendered textures with different camera poses on the photo-realistic simulator. First row: EfficientDetD0; Second row: YOLOv4.}
\label{fig:pose_evaluation}
\end{figure}

\begin{figure*}
\centerline{\includegraphics[width=\textwidth]{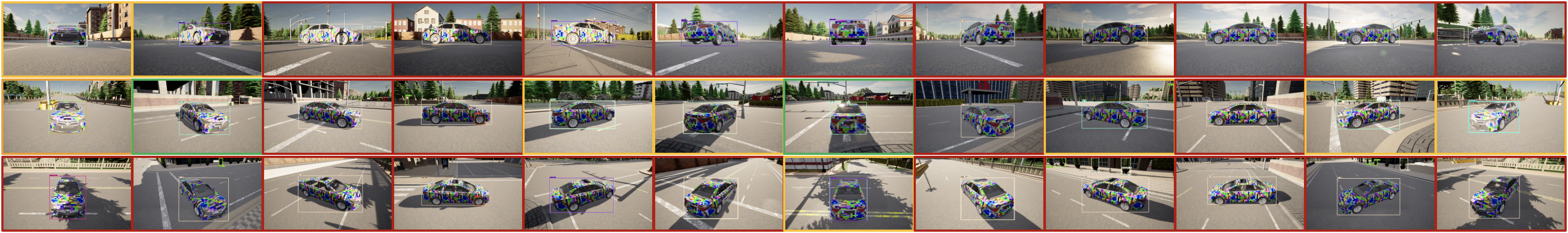}} 
\caption{Adversarial camouflage evaluation on photo-realistic simulator. 
Camera angle is added by 30$^\circ$ for each column, and pitch by 15$^\circ$ for each row.
Border: Red = misdetection; Yellow = partially correct (other labels also detected); Green = correct (detected as car).}
\label{fig:adversarial_pattern_unreal}
\end{figure*}

\begin{table}[]
\caption{Camouflage comparison on photo-realistic simulator. ($\downarrow$) denotes performance drop with respect to normal texture.}
\begin{tabular}{c|lll}
\hline
Model                      & Texture                      & AP@0.5 ($\downarrow$)                     & Conf. ($\downarrow$)                 \\ \hline
\multirow{6}{*}{\specialcell{EffDetD0 \\ \cite{efficientdet}}} & Normal                        & 0.98                       & 0.75                       \\ \cline{2-4} 
                         & Random                       & 0.62 (0.36)                & 0.39 (0.36)                \\ \cline{2-4} 
                           & CAMOU \cite{CAMOU}           & 0.53 (0.44)                & 0.34 (0.41)                \\ \cline{2-4} 
                           & ER \cite{ER}                 & 0.46 (0.51)                & 0.31 (0.47)                \\ \cline{2-4} 
                           & {\textbf{DTA}} & {\textbf{0.34 (0.63)}} & {\textbf{0.27 (0.48)}} \\
                        \hline
\multirow{6}{*}{\specialcell{YOLOv4 \\ \cite{yolov4}}}    & Normal                       & 0.99                       & 0.96                       \\ \cline{2-4} 
                           & Random                       & 0.89 (0.10)                & 0.76 (0.20)                \\ \cline{2-4} 
                           & CAMOU \cite{CAMOU}           & 0.81 (0.18)                & 0.64 (0.32)                \\ \cline{2-4} 
                           & ER \cite{ER}                 & 0.77 (0.22)                & 0.60 (0.35)                \\ \cline{2-4} 
                           & {\textbf{DTA}} & {\textbf{0.76 (0.23)}} & {\textbf{0.57 (0.38)}} \\ \hline
\end{tabular}
\label{table:texture_evaluation}
\end{table}

\textbf{Comparison on Photo-Realistic Simulator} We perform a comparative experiment to evaluate our adversarial camouflage on CARLA simulator. 
We generate evaluation datasets with 200 random locations from four different simulated towns with camera settings: distance (5 m, 10 m, 15 m), pitch angles (0$^\circ$, 15$^\circ$, 30$^\circ$), and a 30 degree rotation interval. This also evaluates whether the generated adversarial camouflage is robust enough to generalize the attack performance on an untrained transformation distribution. 

As shown in Tab. \ref{table:texture_evaluation}, our attack pattern (in bold) lowers both the target AP@0.5 and confidence score more than the random texture or other methods, such as CAMOU and ER, which consider the system as a black box. The details of the average performance for each camera pose can be seen in Fig. \ref{fig:pose_evaluation}, which implies that our adversarial camouflage has the overall lowest AP score on all camera poses.
Additionally, Fig. \ref{fig:adversarial_pattern_unreal} shows the example of our adversarial pattern evaluation with different transformations on the photo-realistic simulator. 
Further evaluation and more samples of the evaluation results can be found in the supplementary material.

\begin{table}[]
\caption{Transferability comparison on photo-realistic simulator.}
\begin{tabular}{l|lll}
\hline
\multirow{3}{*}{\centered{Texture}} & \multicolumn{3}{c}{Average Precision @0.5 ($\downarrow$)}         \\ \cline{2-4} 
                         & \multirow{2}{*}{SSD \cite{ssd}} & Faster & Mask \\ 
                         & & R-CNN \cite{fasterrcnn} & R-CNN \cite{maskrcnn} 
                         \\ \hline
Normal                   & 0.85                 & 0.94                 & 0.94        \\ \hline
Random                   & 0.52 (0.34)          & 0.69 (0.25)          & 0.74 (0.19) \\ \hline
UPC \cite{upcattack}                      & 0.66 (0.20)          & 0.82 (0.11)          & 0.90 (0.03)  \\ \hline
DAS \cite{dasattack}                     & 0.79 (0.06)          & 0.89 (0.04)          & 0.97 (-0.03)\\ \hline
CAMOU \cite{CAMOU}                   & 0.27 (0.59)          & 0.55 (0.39)          & 0.65 (0.29) \\ \hline
ER \cite{ER}                      & 0.27 (0.59)          & 0.56 (0.38)          & 0.64 (0.30) \\ \hline
\textbf{DTA}             & \textbf{0.18 (0.67)} & \textbf{0.41 (0.53)} & \textbf{0.56 (0.37)}   \\ \hline
\end{tabular}

\label{table:transferability_comparison}
\end{table}

We further perform a comparative experiment to evaluate the transferability of our adversarial camouflage to another car (i.e., Audi TT), other transformations, and target models that are not used in the attack phase. We use the same camera settings as those used in the previous experiment. Specifically, UPC and DAS target their original paper's model while the others target EfficientDetD0.

As presented in Tab. \ref{table:transferability_comparison}, our attack also outperforms other camouflage methods in the transferability setting. Notably, UPC and DAS, which are patch-based camouflage methods, show lower performance compared to repeated pattern camouflage, which covers the car's entire paintable surface. Detailed samples as well as a comparison of the methods are included in the supplementary material.

\begin{figure}
\centerline{\includegraphics[width=\columnwidth]{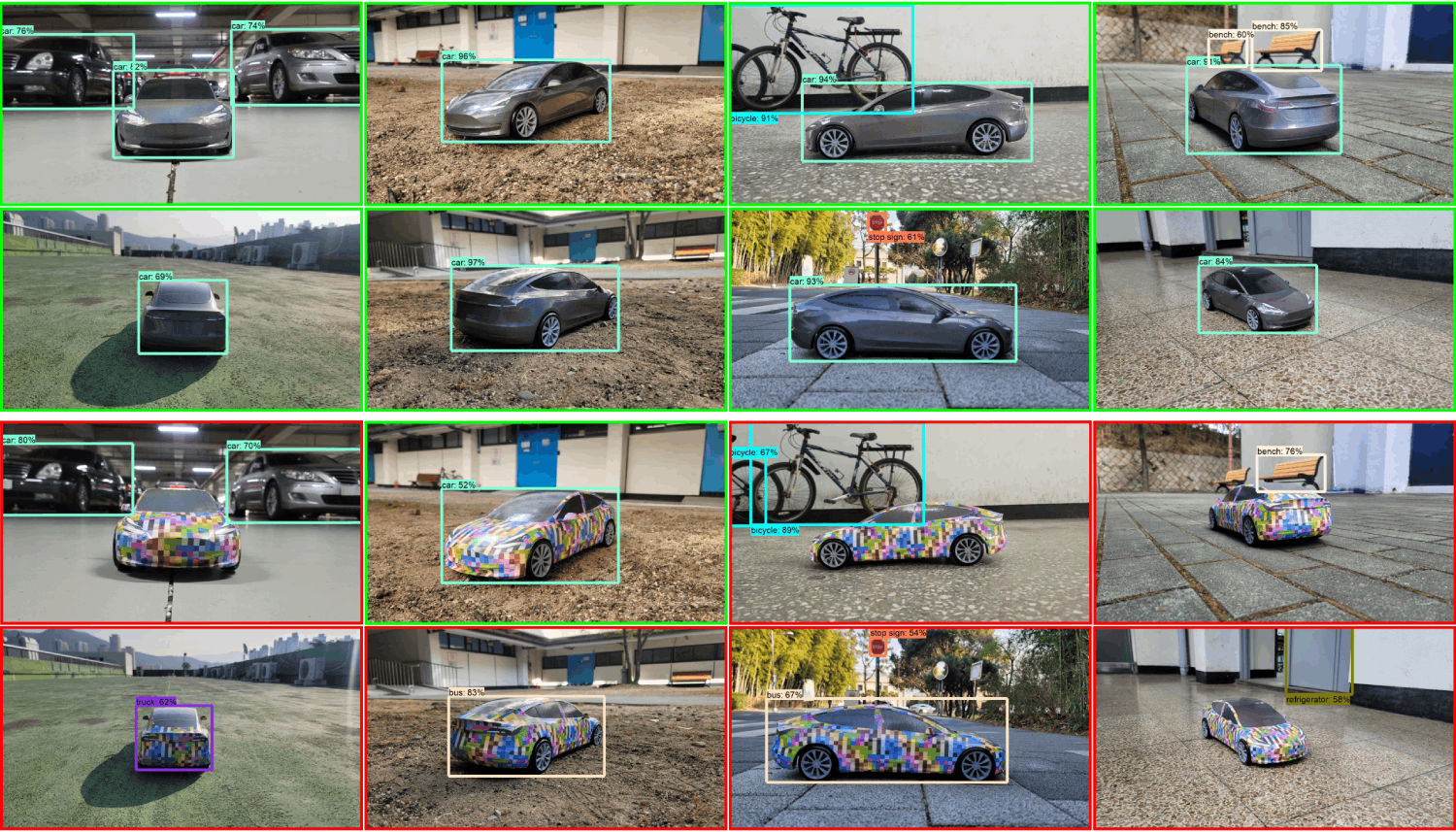}} 
\caption{Real-world evaluation using two 3D-printed scaled models of Tesla Model 3. The upper rows are the standard model while the bottom rows are the attack model.}
\label{fig:real-world-tesla}
\end{figure}

\textbf{Physical Camouflage in Real World} 
We conduct a real-world experiment by manufacturing a 1:10 scaled Tesla Model 3 cars using a 3D printer. Due to limited manufacturing time and resources, we only fabricate two scaled models: one for the camouflaged vehicle with our DTA texture targeting EfficientDetD0, and another for representing a standard vehicle for reference. To achieve a realistic experiment, we place the scaled vehicles in real-life locations indoors and outdoors. We randomly select ten places and capture car images for every 45-degree interval with a similar distance and pitch using a Samsung Galaxy Note 20 Ultra. Figure \ref{fig:real-world-tesla} illustrates how our camouflage can hide the car from the object detection model or result in misdetection while in contrast, the standard vehicle and the background objects are detected correctly. 
On evaluation, results on EfficientDetD0 and YOLOv4 for each vehicle model are presented on Tab. \ref{table:texture_evaluation_real_world}, which shows that our generated camouflage can successfully perform adversarial attack even in the real world.

\begin{table}
\caption{Texture evaluation of vehicle model on real world.}
\begin{tabular}{p{2.25cm}|lll}
\hline
Model                      & Texture                      & AP@0.5 ($\downarrow$)                     & Conf. ($\downarrow$)                 \\ \hline
\multirow{2}{*}{EffDetD0 \cite{efficientdet}} & Normal                       & 0.94                        & 0.83                       \\ \cline{2-4} 
                           & \textbf{DTA}      & \textbf{0.34 (0.60)}        & \textbf{0.35 (0.48)}     \\ \hline
\multirow{2}{*}{YOLOv4 \cite{yolov4}}    & Normal                       & 0.96                       & 0.94                       \\ \cline{2-4} 
                           & \textbf{DTA}       & \textbf{0.61 (0.35)}       & \textbf{0.53 (0.41)}  \\ \hline
\end{tabular}
\label{table:texture_evaluation_real_world}
\end{table}


\section{Discussion}
\textbf{Implications} DTN as a rendering method can easily be extended for use in texture transfer, e.g., \cite{texture_transfer_shape, unsupervised_texture_transfer, inter_texture_transfer}, giving it potentially wide applicability, and it is relatively harmless to the public. 
On the other hand, DTA as an attack certainly poses potentially dangerous consequences
if it is to be implemented. Firstly, the attack pattern is more robust, as it accounts for the photo-realistic effect, leading to an increase in its attack success rate. This is concerning, particularly in the future era of autonomous vehicles. Secondly, as pointed out in \cite{CAMOU}, the method of printing the adversarial pattern on vehicles rather than on traffic signs also adds to its hazard since painting cars can be done legally, whereas the latter is a violation and more likely to be removed by the authorities. On cars, it would make it easier for the adversary to launch the attack, and it could even be done in a coordinated fashion with other adversaries' cars. To mitigate this attack, one possible solution is to reinforce object detection models with adversarial training \cite{pgd, ensemble}.

\textbf{Limitations} 
DTN leverages a simple projection instead of a more sophisticated mapping for transferring the texture to the target object. Hence, an inaccurate texture may be produced for objects with a complex shape. For mitigation, we currently generate the pattern with random scaling, shifting, and rotation to improve robustness. 
Additionally, the current approach has not considered the naturalness of the pattern so that it does not look suspicious to a person, which may be important for some cases. This can be done by implementing additional loss, such as the smooth loss introduced in \cite{accessorize_to_crime}. In the future, we plan to investigate further the internal texture mapping and naturalness of our adversarial camouflage approach.

\section{Conclusion}
In this paper, we proposed DTA, a framework that considers differentiability as well as photo-realistic aspects in its adversarial pattern generation, giving robust adversarial camouflage at any viewing angle. In particular, we leveraged our novel rendering technique, namely DTN, which can extract the expected transformation of a rendered object and retain its original attributes. Our experiments included a comparison with previous works (i.e., UPC \cite{upcattack}, DAS \cite{dasattack}, CAMOU \cite{CAMOU}, and ER \cite{ER}) in a photo-realistic simulator as well as a demonstration in the real world, showing the applicability and transferability of our approach. 

\section{Acknowledgments}
This work was supported by Institute of Information \& Communications Technology Planning \& Evaluation (IITP) grant funded by the Korea government (MSIT) (No.2019-0-01343, Regional strategic industry convergence security core talent training business), and by a grant from Defense Acquisition Program Administration and Agency for Defense Development under contract UE201131RD.

{\small
\bibliographystyle{ieee_fullname}
\bibliography{main}
}

\newpage
\onecolumn
\appendix
\appendixpage
{\Large\centering \textbf{Supplementary Material}}


\section{Overview}
In this supplementary material, we describe our detailed algorithm, experiment setup, evaluation details, and evaluation results that can not be included in the main paper due to limited space. Furthermore, we also provide additional experiments and more evaluation samples in both simulated environments and real world.

\section{Algorithm Details}
\begin{figure*}
    \centering
    \includegraphics[width=0.95\textwidth]{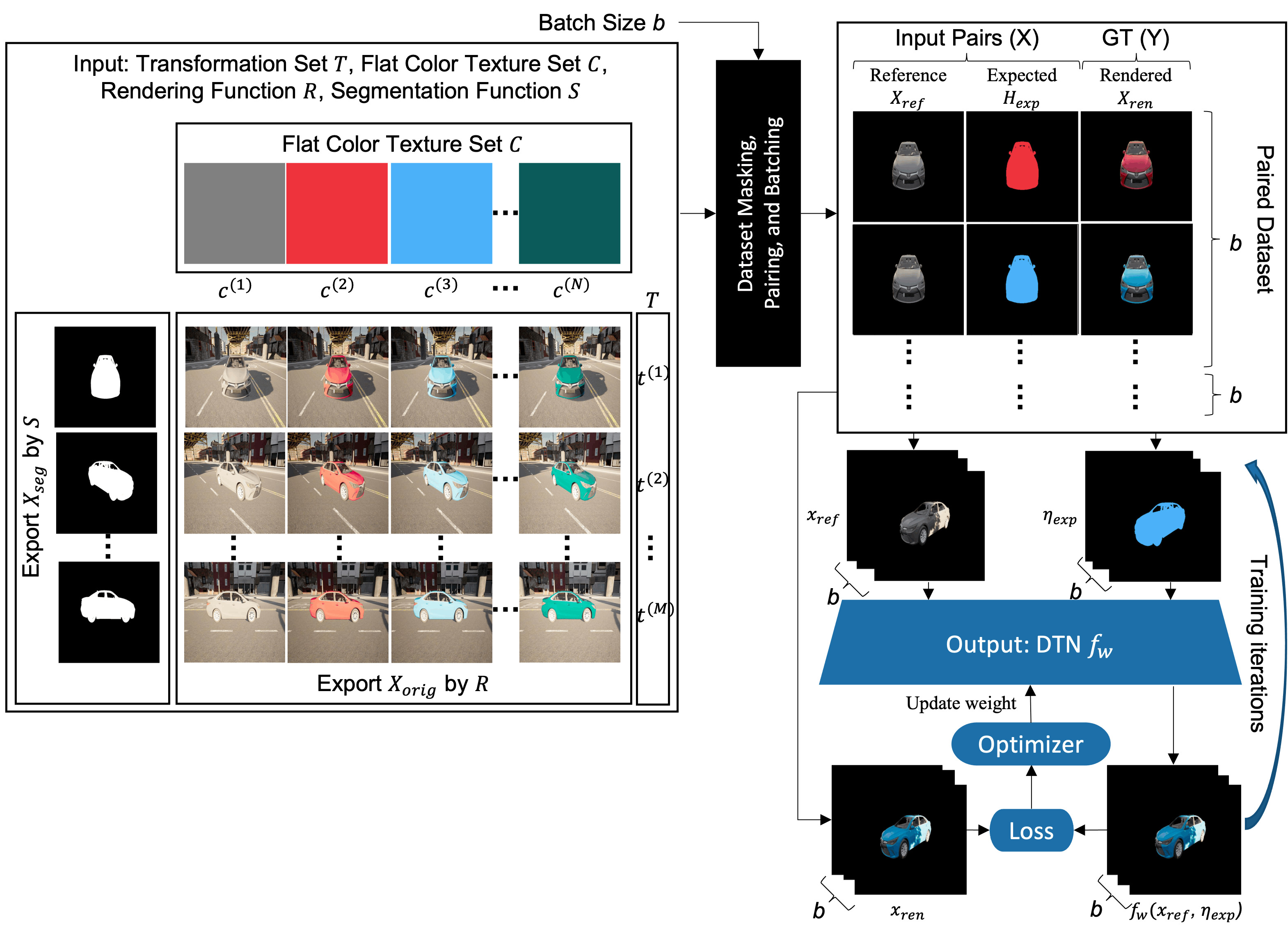}
    \caption{Training process of DTN.}
    \label{fig:train_dtn}
\end{figure*}

In this section, we provide detailed algorithm and figures about the training process of the Differentiable Transformation Network (DTN), our proposed texture rendering method. Algorithm \ref{alg:train_dtn} and Figure \ref{fig:train_dtn} describe the detailed training process of the DTN.

\begin{algorithm}[]
    \caption{Training process of the DTN.}
    \label{alg:train_dtn}
\begin{algorithmic}
    \STATE {\bfseries Input:} Transformation set $T = \{t^{(1)},\dots,t^{(M)}\}$, Flat color texture set $C = \{c^{(1)},\dots,c^{(N)}\}$, Rendering function $R$, Segmentation function $S$
    \STATE {\bfseries Output:} DTN $f_w$
    \STATE (1) Export $X_{orig}$ and $X_{seg}$ from the photo-realistic rendering engine
    \FOR {$m = 1$ to $M$}
    \STATE $X_{seg}^{(m)} \gets S(t^{(m)})$
    \FOR {$n = 1$ to $N$}
    \STATE $X_{orig}^{(m)(n)} \gets R(t^{(m)}, c^{(n)})$
    \ENDFOR
    \ENDFOR
    \STATE (2) Dataset masking and pairing
    \FOR {$m = 1$ to $M$}
    \FOR {$n = 1$ to $N$}
    \STATE $X_{ref} \gets X_{ref} \cup (X_{orig}^{(m)(1)} \times X_{seg}^{(m)})$
    \STATE $H_{exp} \gets H_{exp} \cup (c^{(n)} \times X_{seg}^{(m)})$
    \STATE $X_{ren} \gets X_{ren} \cup (X_{orig}^{(m)(n)} \times X_{seg}^{(m)})$
    \ENDFOR
    \ENDFOR
    \STATE (3) DTN Training
    \STATE Initialize $f_w$ with random weights $w$
    \FOR {number of training iterations}
    \STATE Sample minibatch of each $b$ samples $x_{ref} \in X_{ref}$, $\eta_{exp} \in H_{exp}$, $x_{ren} \in X_{ren}$
    \STATE Update $f_w$ with gradient descent based on the binary cross-entropy loss between $f_w(x_{ref}, \eta_{exp})$ and $x_{ren}$
    \ENDFOR
\end{algorithmic}
\end{algorithm}

\section{Experiment Details}

\subsection{Dataset Details}
\textbf{For DTN Training} We generate the datasets used for training and evaluating DTN model using CARLA Simulator. We use Toyota Camry as our target object, i.e., where the adversarial pattern will be generated and evaluated. We select a map (Town03 in CARLA) and randomly choose 75 spawn locations to generate both training and validation datasets, and choose 50 spawn locations for generating testing datasets. For every location, we spawn the target car and capture the photo realistic images using cameras with 5 m distance, 15-degree pitch, and for every 45-degree rotation. For each spawn location and camera pose, we sequentially change the car texture with 50 random flat color textures and get a segmentation image. The segmentation image is used for extracting either target object or background as needed by DTN. Datasets used for training and testing use different spawn locations and different random color textures. The datasets are then grouped by the same transformation and same texture as shown in Figure \ref{fig:train_dtn}.

\textbf{For Adversarial Camouflage Generation}
We select 250 spawn locations in the same map and generate datasets of the target object with the flat color texture used as a reference image during DTN training. The camera pose used for adversarial pattern generation is the same camera pose used during DTN training.

\begin{figure}
    \centering
    \begin{subfigure}[b]{0.45\columnwidth}
        \centering
        \includegraphics[width=0.56\columnwidth]{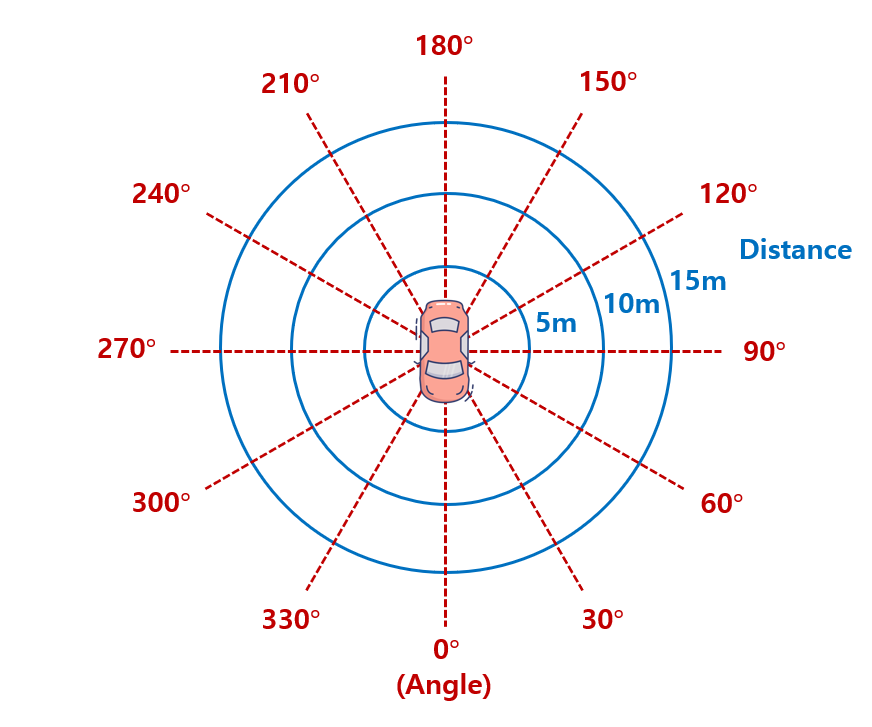}
        \caption{Various camera angle}
    \end{subfigure}
    \begin{subfigure}[b]{0.45\columnwidth}
        \centering
        \includegraphics[width=0.56\columnwidth]{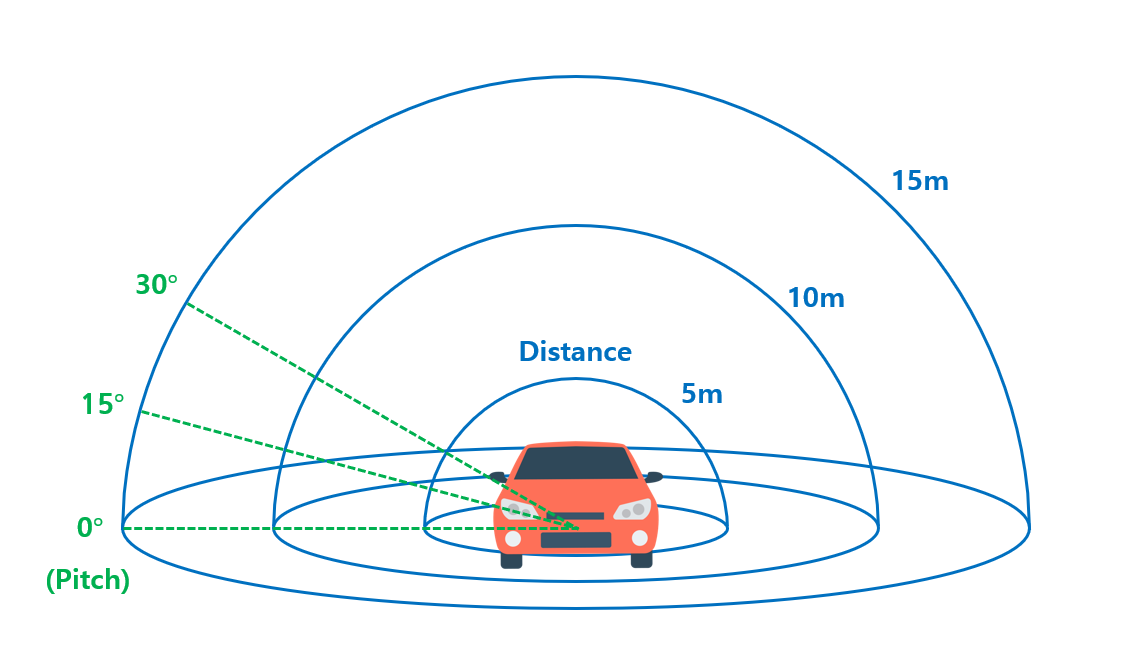}
        \caption{Various camera pitch}
    \end{subfigure}
    \caption{Camera pose distribution for adversarial camouflage evaluation.}
    \label{fig:camera_pose_distributiion}
\end{figure}

\textbf{For Adversarial Pattern Evaluation}
We use more camera poses and different towns to evaluate whether the generated attack pattern is robust enough to generalize the attack performance on untrained transformation distribution. 
We selected 200 spawn locations from four simulated towns (Town01, Town04, Town05, and Town06 in CARLA) with minimum occlusions. For each location, we spawn the textured car and capture images with camera settings: distance (5 m, 10 m, 15 m), pitch angles (0$^\circ$, 15$^\circ$, 30$^\circ$), and a 30-degree rotation interval as described in Figure \ref{fig:camera_pose_distributiion}.

\subsection{Implementation Details of Compared Methods}

We provide detailed parameter setup of compared methods: CAMOU \cite{CAMOU}, ER \cite{ER}, UPC \cite{upcattack} and DAS \cite{dasattack} in our experiment.

\textbf{Repeated Pattern-based Physical Camouflage}
CAMOU and ER produce the repeated patterns covering all target object surfaces as the physical camouflage output. We select the same $16 \times 16$ attack resolution following the best CAMOU result to conduct a fair comparison. We closely follow the approach to replicate the original papers, but we rebuild the environment and target models based on our evaluation setup (Generating attack pattern on Toyota Camry and targeting both EfficientDetD0 and YOLOv4). 
For CAMOU, we use the same clone network architecture and other parameters as the original paper for generating the camouflage pattern.
For ER, we also use the same parameters as the original paper, except that we change $p = 3$ and $r = 1$ parameters so that it outputs a same $16 \times 16$ texture.

\begin{figure}[h]
    \centering
    \begin{subfigure}[b]{0.49\columnwidth}
        \centering
        \includegraphics[width=0.9\columnwidth]{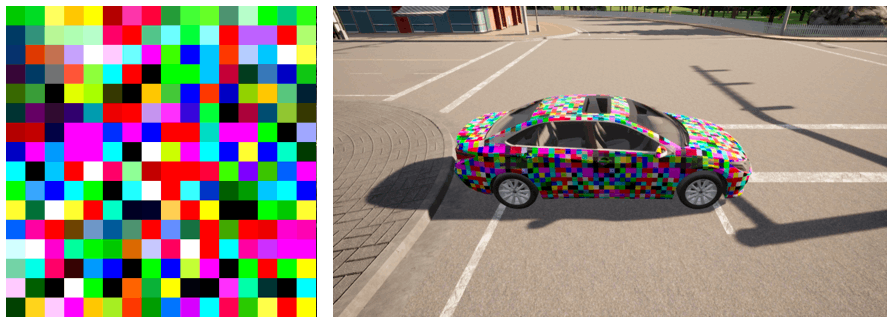}
        \caption{CAMOU \cite{CAMOU}}
    \end{subfigure}
    \begin{subfigure}[b]{0.49\columnwidth}
        \centering
        \includegraphics[width=0.9\columnwidth]{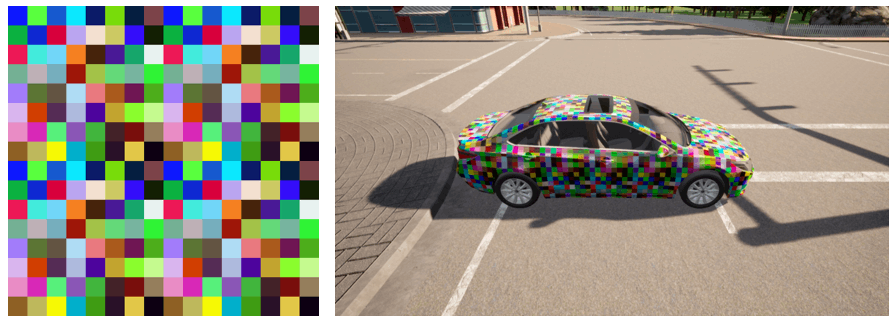}
        \caption{Enlarge and Repeat (ER) \cite{ER}}
    \end{subfigure}
    \caption{Repeated pattern-based physical camouflage from original CAMOU \cite{CAMOU} and ER \cite{ER} when re-implemented in our environment. For each Figure (a) and (b), left side shows the generated camouflage pattern, while the right side is the result of a car rendered by repeated texture using world-aligned texture.}
    \label{fig:upc_das_implementation}
\end{figure}

\textbf{Patch-based Physical Camouflage}
UPC and DAS produce patches that are attached to specific target object surface as the physical camouflage. Since they have different settings to recreate in our environment, we only evaluate them on the transferability experiment.
We run their official code and keep the original attack setup on cars, and then extract their attack patches to our environment. 
We use decals in Unreal for painting their patches to our car surfaces.
To conduct a fair comparison, we paint the patches into car hood, doors, rooftop, and back so that they can be seen from various angles. Figure \ref{fig:upc_das_implementation} shows the sample of how we re-implement the generated patches to our transferability test environment.

\begin{figure}[h]
    \centering
    \begin{subfigure}[b]{0.49\columnwidth}
        \centering
        \includegraphics[width=0.9\columnwidth]{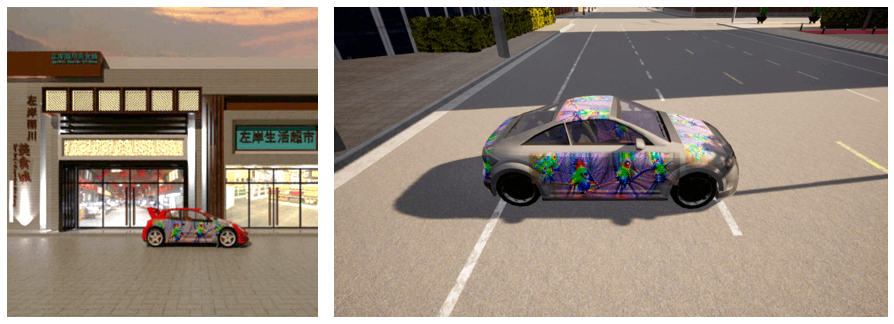}
        \caption{Universal Physical Camouflage (UPC) \cite{upcattack}}
    \end{subfigure}
    \begin{subfigure}[b]{0.49\columnwidth}
        \centering
        \includegraphics[width=0.9\columnwidth]{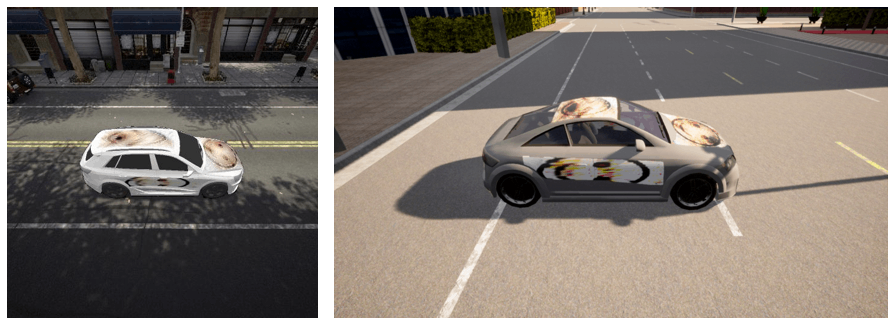}
        \caption{Dual Attention Suppression (DAS) Attack \cite{dasattack}}
    \end{subfigure}
     \caption{Transferability: implementation of original camouflage of UPC \cite{upcattack} and DAS \cite{dasattack} in our environment. For each (a) and (b), the left and right side shows the sample of the original paper and in our test environment, respectively.}
    \label{fig:upc_das_implementation}
\end{figure}

\section{Evaluation Details}
\subsection{DTN Evaluation}
\begin{figure}[h]
    \centering
    \begin{subfigure}[b]{0.49\textwidth}
        \centering
        \includegraphics[width=0.85\textwidth]{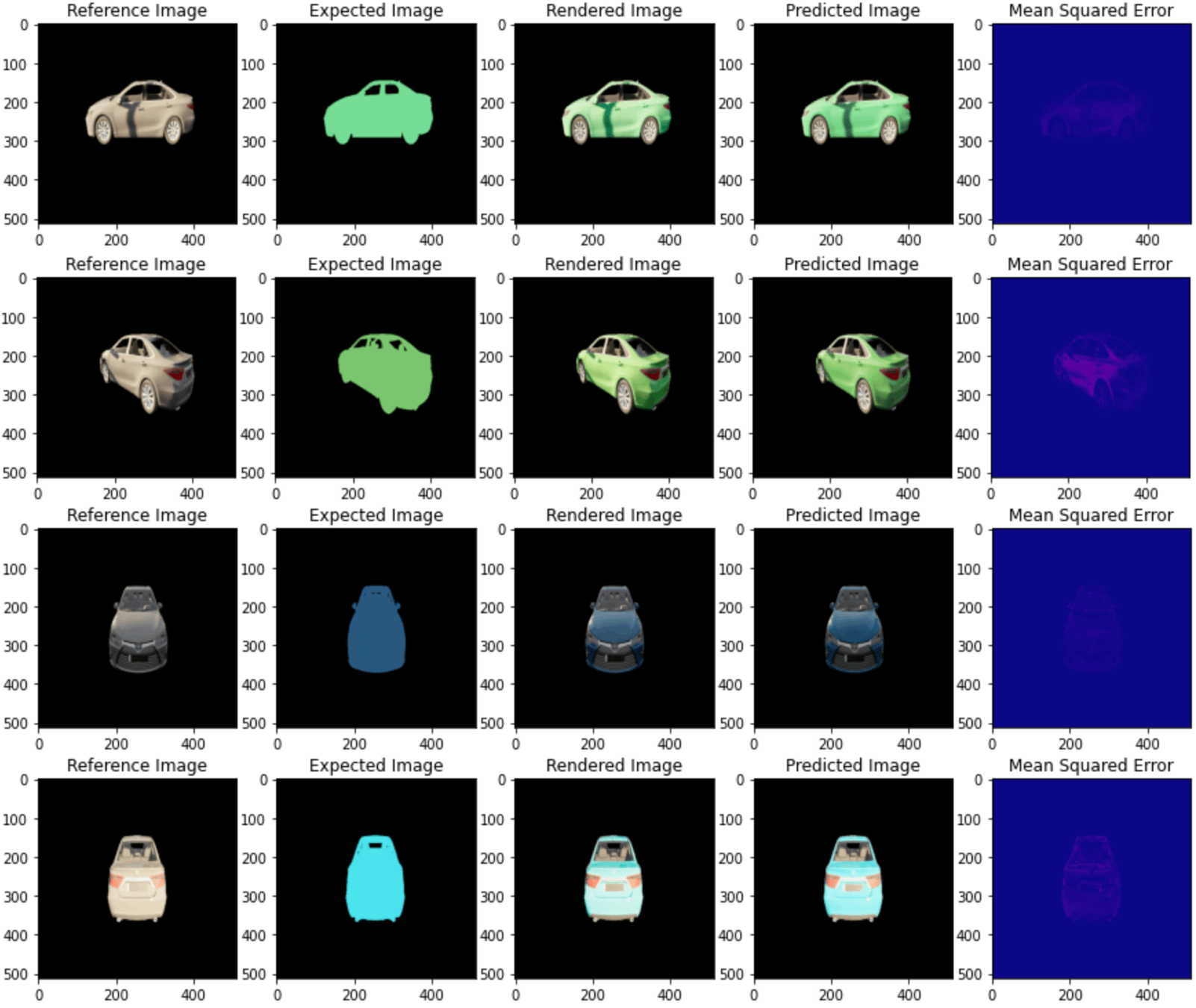}
        \caption{Prediction on Validation set}
    \end{subfigure}
    \begin{subfigure}[b]{0.49\textwidth}
        \centering
        \includegraphics[width=0.85\textwidth]{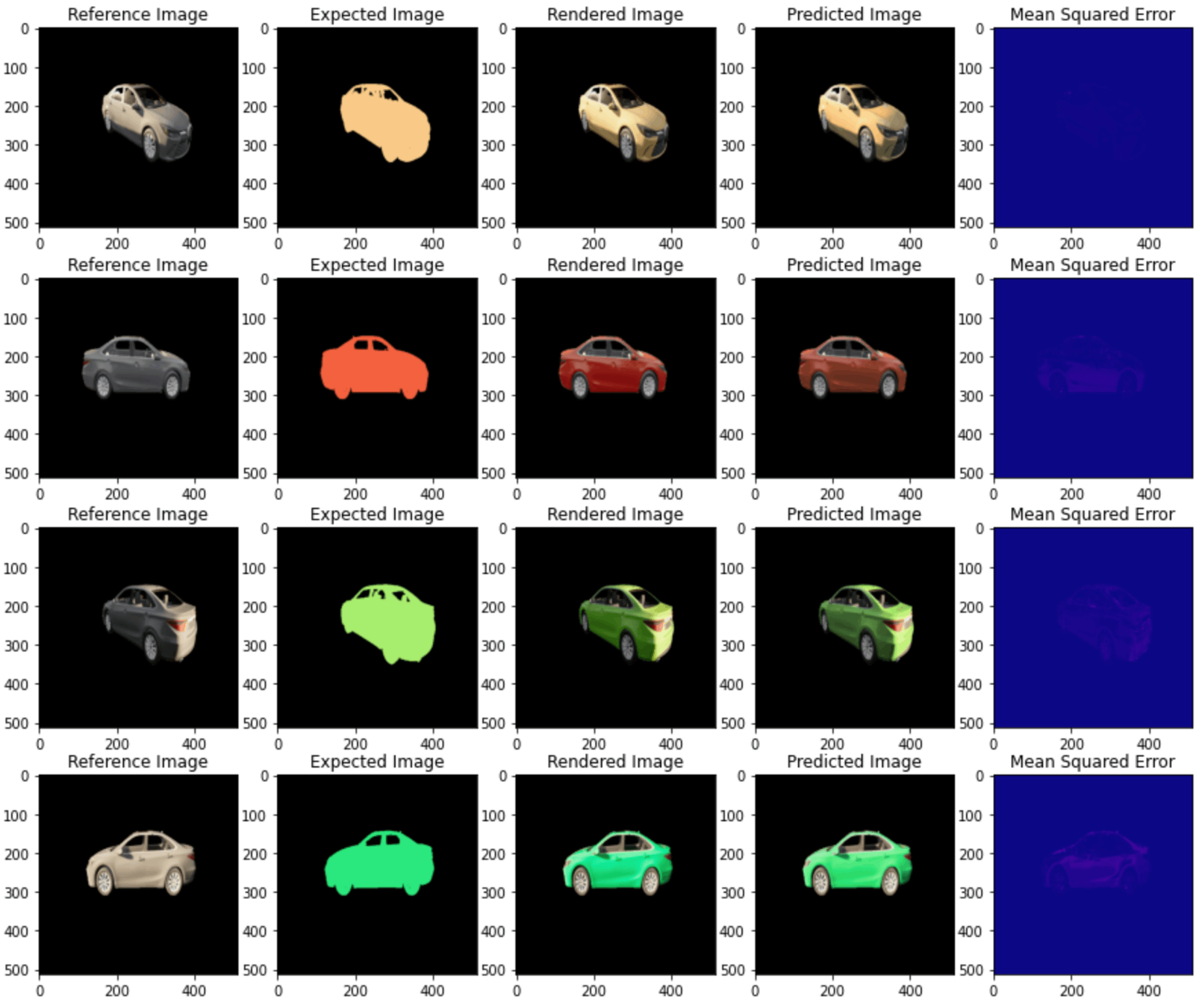}
        \caption{Prediction on Test set}
    \end{subfigure}
    \caption{DTA sample predictions on unseen data}
    \label{fig:eotnet_prediction}
\end{figure}

\textbf{Prediction results}
We visualize the sample prediction results of DTA on unseen data after the training is complete. Figure \ref{fig:eotnet_prediction} shows sample prediction result of trained DTA on validation and test datasets. As shown, the model can learn how to render the expected image given the inputs. Apart from learning the texture color changes caused by object material and lighting, DTA can also extract parts that do not need to be changed from reference images, such as tires, car headlights, interiors, license plate, etc. This provides convenience to the user without the need to separate the masking from the part where the texture will be applied.

\begin{figure}[h]
    \centering
    \begin{subfigure}[b]{0.49\textwidth}
        \centering
        \includegraphics[width=0.85\textwidth]{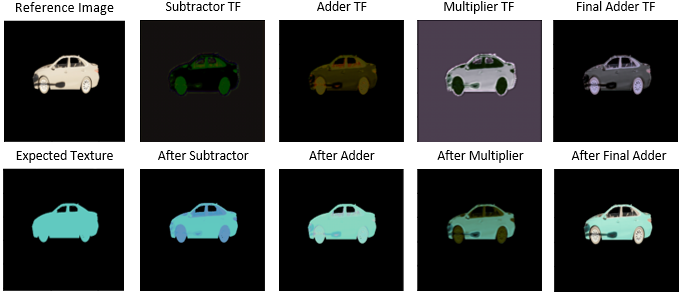}
    \end{subfigure}
    \begin{subfigure}[b]{0.49\textwidth}
        \centering
        \includegraphics[width=0.85\textwidth]{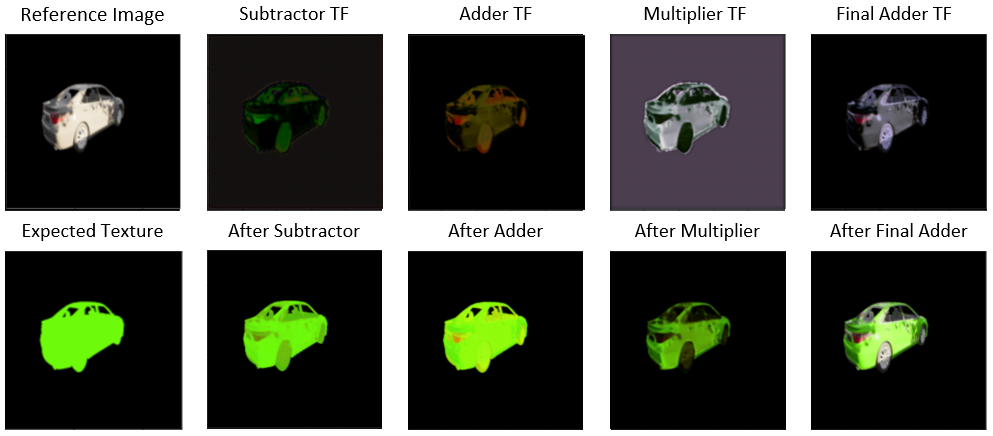}
    \end{subfigure}
    \begin{subfigure}[b]{0.49\textwidth}
        \centering
        \includegraphics[width=0.85\textwidth]{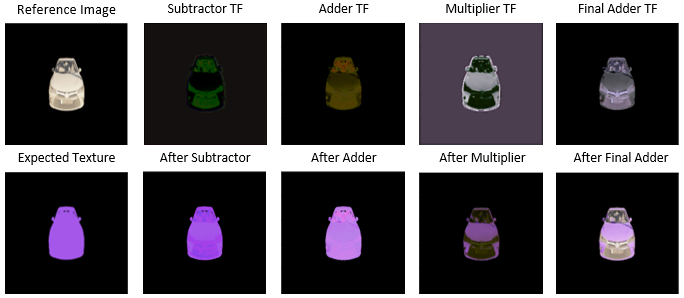}
    \end{subfigure}
    \begin{subfigure}[b]{0.49\textwidth}
        \centering
        \includegraphics[width=0.85\textwidth]{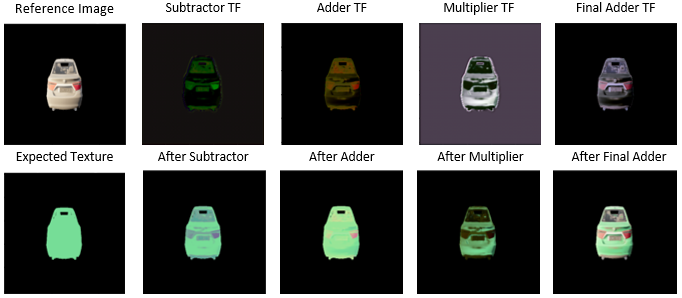}
    \end{subfigure}
    \caption{Visualization of each transformation process.}
    \label{fig:eotnet_transformation_visualization}
\end{figure}

\textbf{Visualizing Transformation Features} We visualize the output of each encoded transformation features $TF$ and how it is used to transform the expected texture $\eta_{exp}$.
On Fig. \ref{fig:eotnet_transformation_visualization}, it can be observed that the first subtractor and adder $TF$ encode the color transformation including the part where the texture is not applied (e.g., wheels, interior, etc). $TF$ multiplier encodes how the texture is minimized or emphasized. Since the image is visualized in RGB format, white or gray color indicate that the proportion between each color channel is similar. Additionally, $TF$ multiplier also removes the nonapplied texture part. Finally, the final $TF$ adder gives the final touch for transforming the expected texture to the rendered version, which includes adding the nonapplied texture part from the reference image. 

\begin{figure}[h]
    \centering
    \begin{subfigure}[b]{\textwidth}
        \centering
        \includegraphics[width=\textwidth]{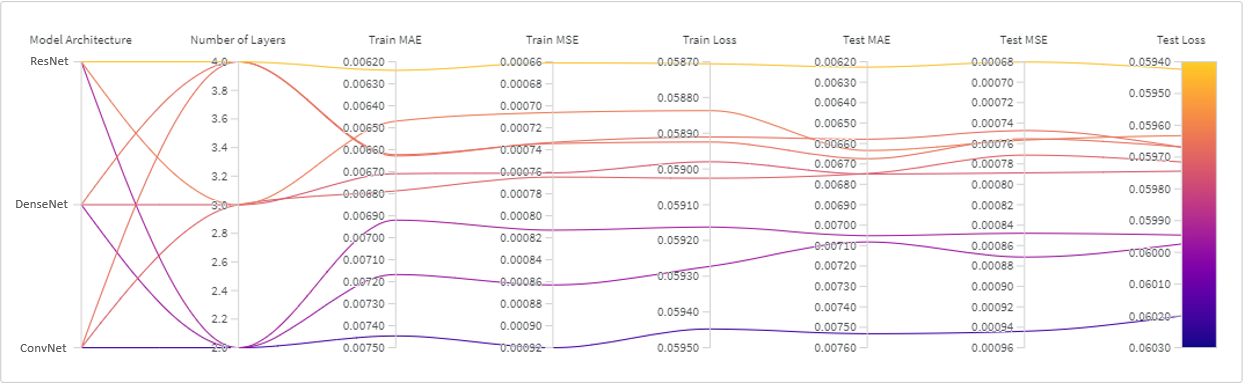}
        \caption{Parallel coordinates charts for evaluating effect of different model architecture and number of layers with respect to evaluation metrics.}
        \label{fig:eotnet_parallel_coordinate}
    \end{subfigure}
    \begin{subfigure}[b]{0.49\textwidth}
        \centering
        \includegraphics[width=\textwidth]{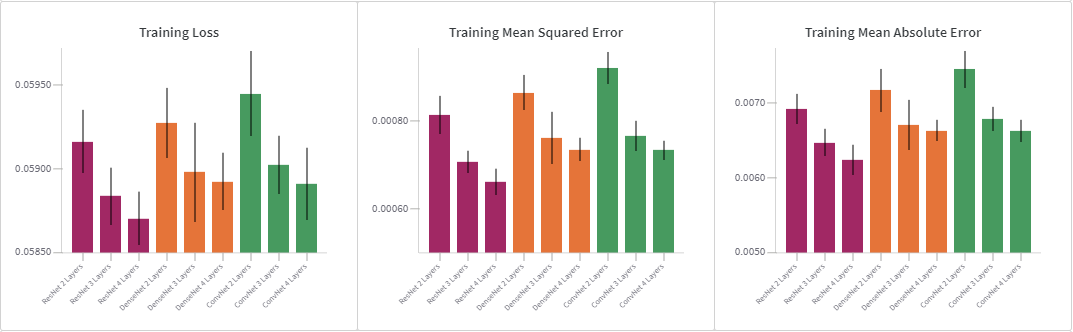}
        \caption{Training Loss, Mean Squared Error, and Mean Absolute Error.}
        \label{fig:eotnet_train_eval}
    \end{subfigure}
    \begin{subfigure}[b]{0.49\textwidth}
        \centering
        \includegraphics[width=\textwidth]{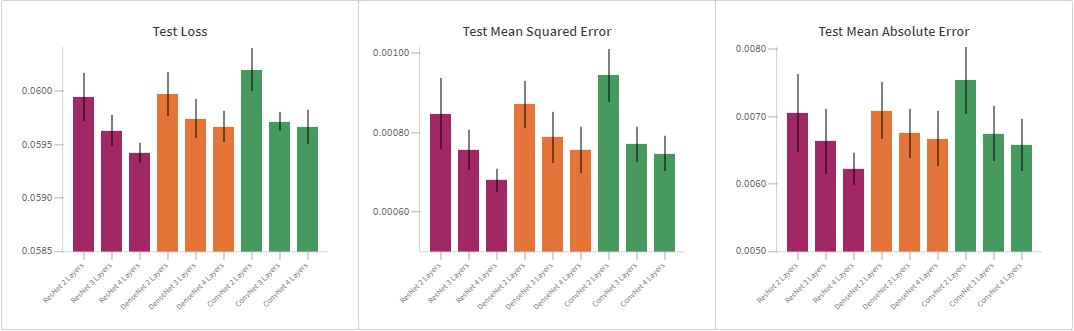}
        \caption{Test Loss, Mean Squared Error, and Mean Absolute Error.}
        \label{fig:eotnet_test_eval}
    \end{subfigure}
    \begin{subfigure}[b]{0.49\textwidth}
        \centering
        \includegraphics[width=\textwidth]{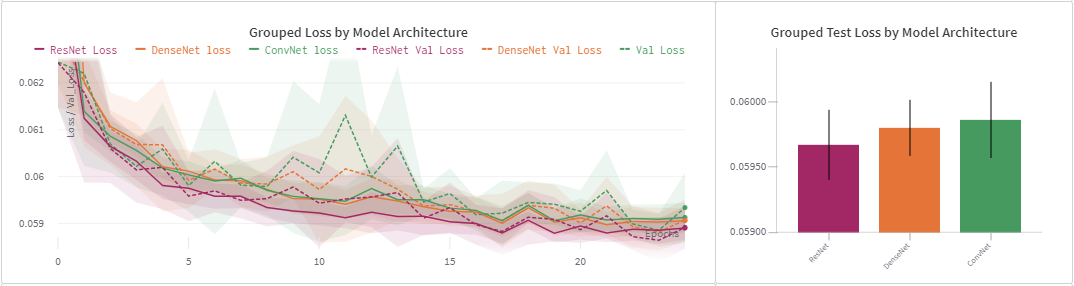}
        \caption{Grouped Loss by Model Architecture.}
        \label{fig:eotnet_grouped_loss_by_arch}
    \end{subfigure}
    \begin{subfigure}[b]{0.49\textwidth}
        \centering
        \includegraphics[width=\textwidth]{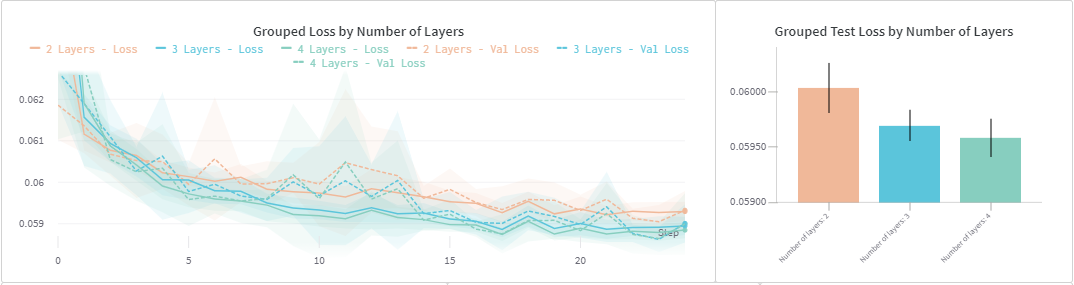}
        \caption{Grouped Loss by Number of Layers $k$.}
        \label{fig:eotnet_grouped_loss_by_number_of_layers}
    \end{subfigure}
    \caption{DTN evaluation with different architectures [ResNet (red), DenseNet (orange), ConvNet (green)] and number of layers $k=[2, 3, 4]$ .}
    \label{fig:eotnet_architecture_evaluation}
\end{figure}

\pagebreak
\textbf{Different Architectures}
We further evaluate and visualize the performance of DTN with different architectures [ResNet, DenseNet, Plain CNN] and $k = [2, 3, 4]$ layers. We run the evaluation five times and use the same unique random seeds for each run on different model.
The evaluation results can be seen in Figure \ref{fig:eotnet_architecture_evaluation}.

Figure \ref{fig:eotnet_parallel_coordinate} shows parallel coordinate chart for evaluating the effect of different architectures (first column) and number of layers (second column) with respect to train and test evaluation (i.e., Mean Absolute Error, Mean Squared Error, and Loss, respectively). The values are the mean over multiple runs. All evaluation metrics columns are sorted where brighter colors gradient (yellow) are better.

Figure \ref{fig:eotnet_train_eval} and \ref{fig:eotnet_test_eval} show the detailed training and test evaluation bar plot for every architecture (ResNet [Red], DenseNet [Orange], and Plain CNN [Green]) with different number of layers $k$. The bar contains the mean and standard deviation information over multiple runs.

Figure \ref{fig:eotnet_grouped_loss_by_arch} and \ref{fig:eotnet_grouped_loss_by_number_of_layers} describe grouped loss, including train/validation loss for each epochs and final test loss. The difference is whether the loss is grouped by architecture or number of layers. This shows the effect of each parameter on the evaluation metric more clearly.

\textbf{Different Reference Types.}
We select the best-performing model and re-evaluate DTN by varying the training configuration such as whether to use single fixed or multiple flat color texture for the reference input.

\begin{figure}[h]
    \centering
    \begin{subfigure}[b]{\textwidth}
        \centering
        \includegraphics[width=\textwidth]{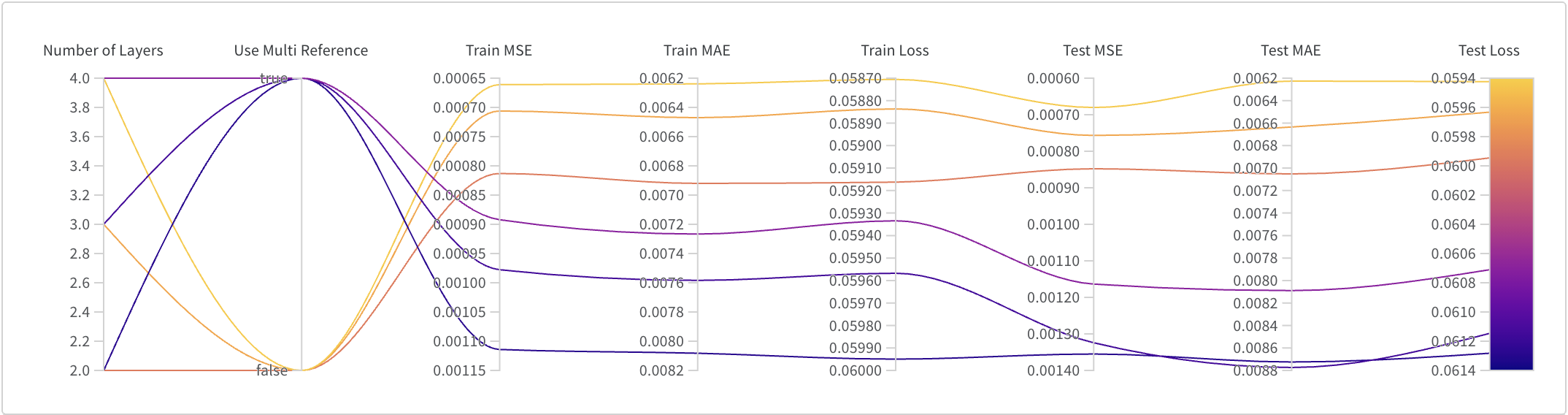}
        \caption{Parallel coordinates charts for evaluating effect of different reference type and number of layers respected to evaluation metrics.}
        \label{fig:multiref_parallel_coordinate}
    \end{subfigure}
    \begin{subfigure}[b]{0.49\textwidth}
        \centering
        \includegraphics[width=\textwidth]{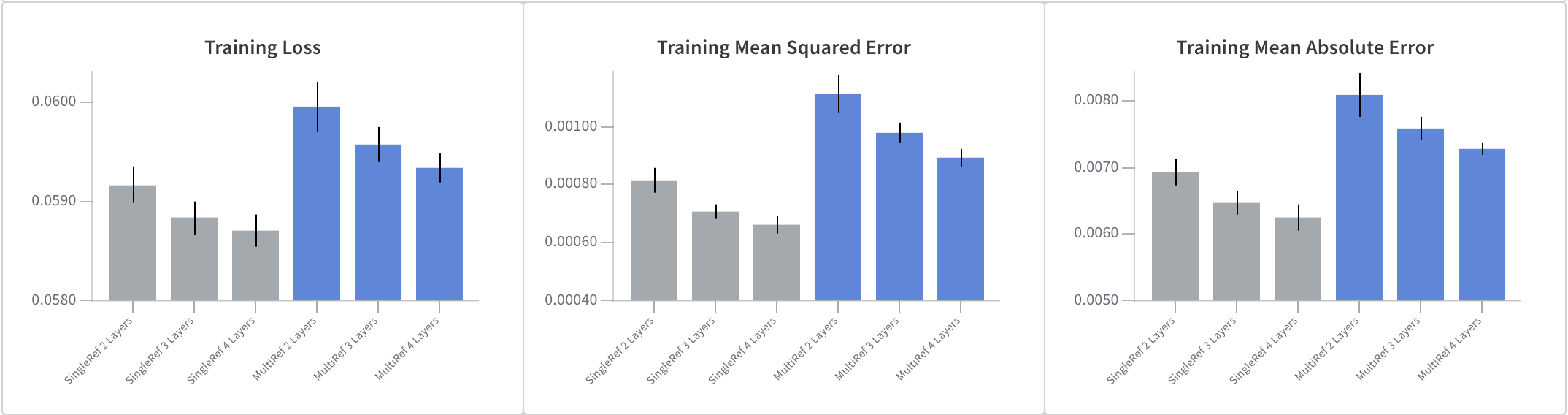}
        \caption{Training Loss, Mean Squared Error, and Mean Absolute Error.}
        \label{fig:eotnet_train_eval_ref}
    \end{subfigure}
    \begin{subfigure}[b]{0.49\textwidth}
        \centering
        \includegraphics[width=\textwidth]{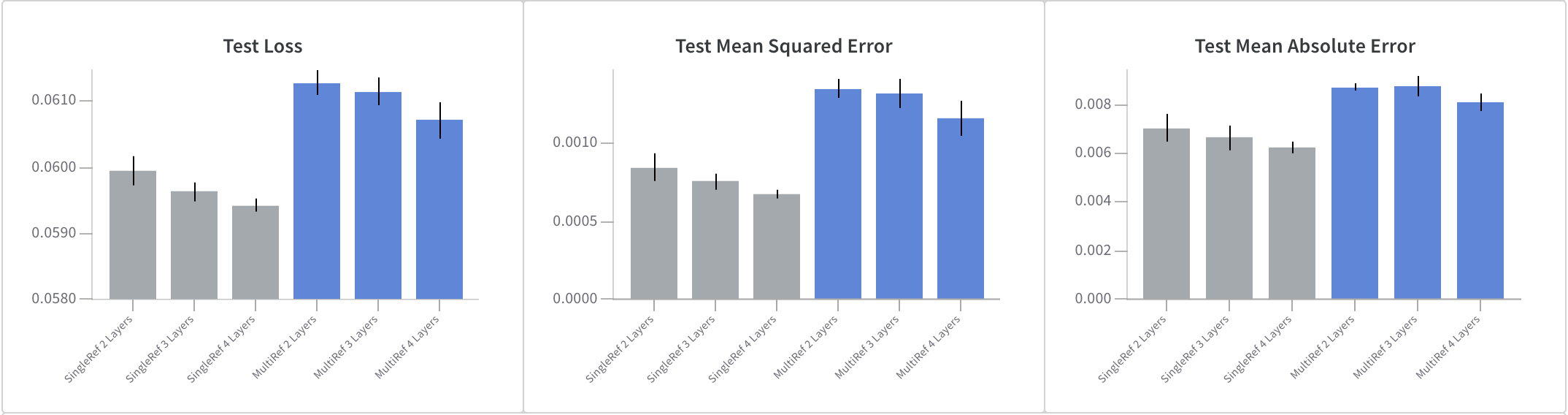}
        \caption{Test Loss, Mean Squared Error, and Mean Absolute Error.}
        \label{fig:eotnet_test_eval_ref}
    \end{subfigure}
    \begin{subfigure}[b]{0.49\textwidth}
        \centering
        \includegraphics[width=\textwidth]{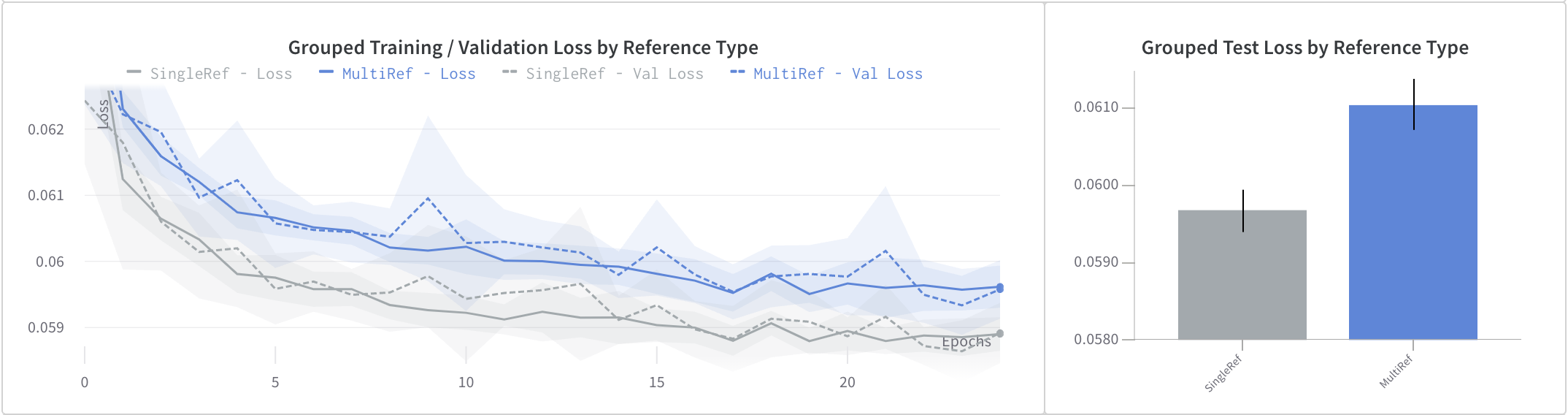}
        \caption{Grouped Loss by Reference Types.}
        \label{fig:eotnet_grouped_loss_by_ref}
    \end{subfigure}
    \begin{subfigure}[b]{0.49\textwidth}
        \centering
        \includegraphics[width=\textwidth]{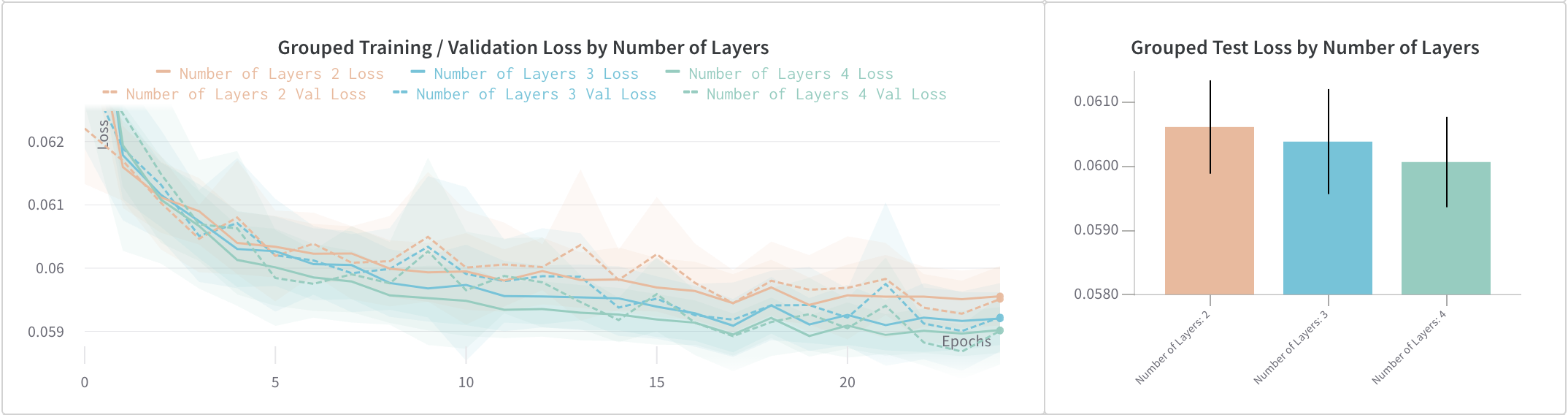}
        \caption{Grouped Loss by Number of Layers $k$.}
        \label{fig:eotnet_grouped_loss_by_ref_number_of_layers}
    \end{subfigure}
    \caption{DTN evaluation with different reference types [SingleRef (gray), MultiRef (blue)] and number of layers $k=[2, 3, 4]$.}
    \label{fig:eotnet_ref_evaluation}
\end{figure}

\begin{figure}[h]
    \centering
    \begin{subfigure}[b]{0.49\textwidth}
        \centering
        \includegraphics[width=\textwidth]{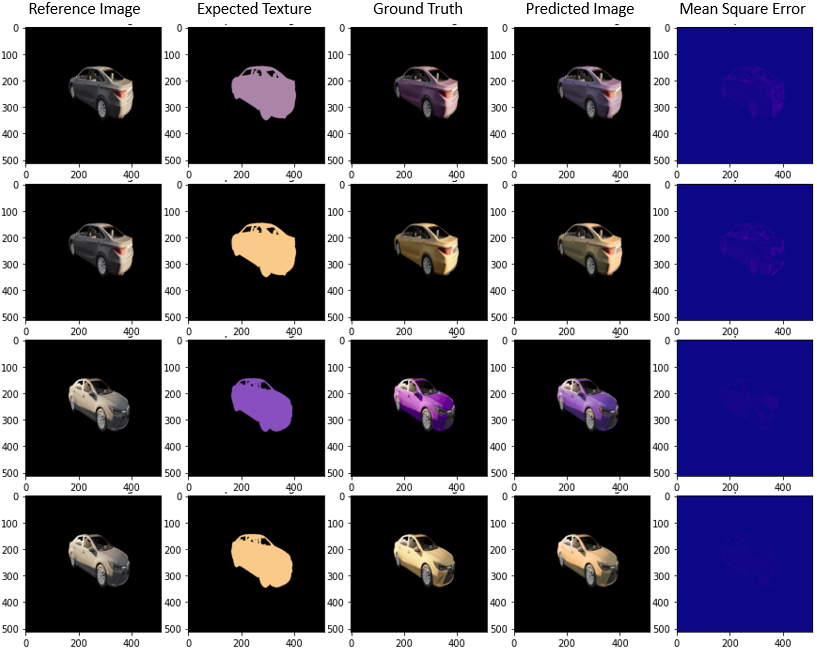}
        \caption{Prediction on Single/Fixed Reference Trained Model}
    \end{subfigure}
    \begin{subfigure}[b]{0.49\textwidth}
        \centering
        \includegraphics[width=\textwidth]{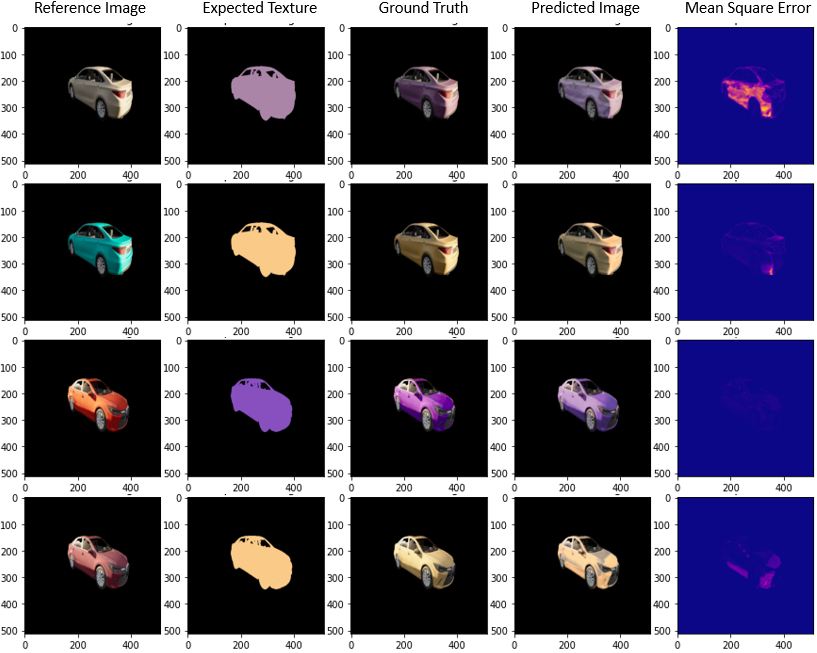}
        \caption{Prediction on Multi Reference Trained Model}
    \end{subfigure}
    \caption{DTN sample predictions for single/fixed-reference vs multi-reference model. Both models have the same parameters except for the type of reference used.}
    \label{fig:eotnet_ref_prediction}
\end{figure}

Figure \ref{fig:eotnet_ref_evaluation} shows the evaluation of DTN with different reference types [SingleRef or MultiRef] and $k = [2, 3, 4]$ layers.
Figure \ref{fig:eotnet_ref_prediction} shows sample prediction result of trained model between single/fixed-reference vs multi-reference.
Overall figures show that training DTN with single-reference texture is better than using multi-reference texture. 
Learning the transformation from a single fixed color texture reference is considered as a simpler task. 
Even though we can increase the number of layers for lowering the loss, our final goal of training DTN is for generating the adversarial pattern. In this case, using a single reference color image as the reference dataset is considered the best practice as it has a lower reconstruction loss.

\hphantom{}

\hphantom{}

\hphantom{}

\pagebreak

\subsection{Adversarial Camouflage Evaluation}
\begin{figure}[h]
    \centering
    \begin{subfigure}[b]{0.49\textwidth}
        \centering
        \includegraphics[width=\textwidth]{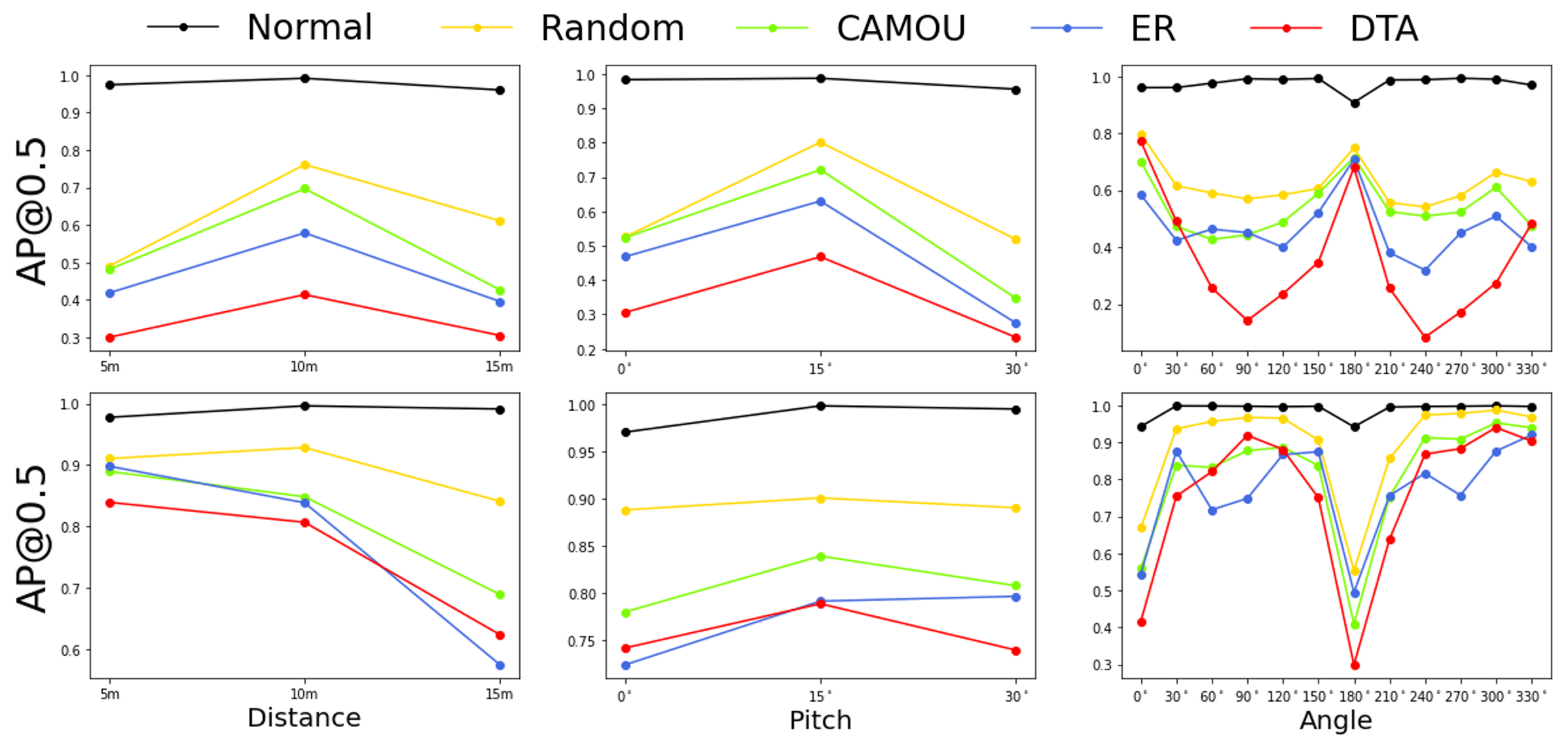}
        \caption{Target Average Precision@0.5}
    \end{subfigure}
    \begin{subfigure}[b]{0.49\textwidth}
        \centering
        \includegraphics[width=\textwidth]{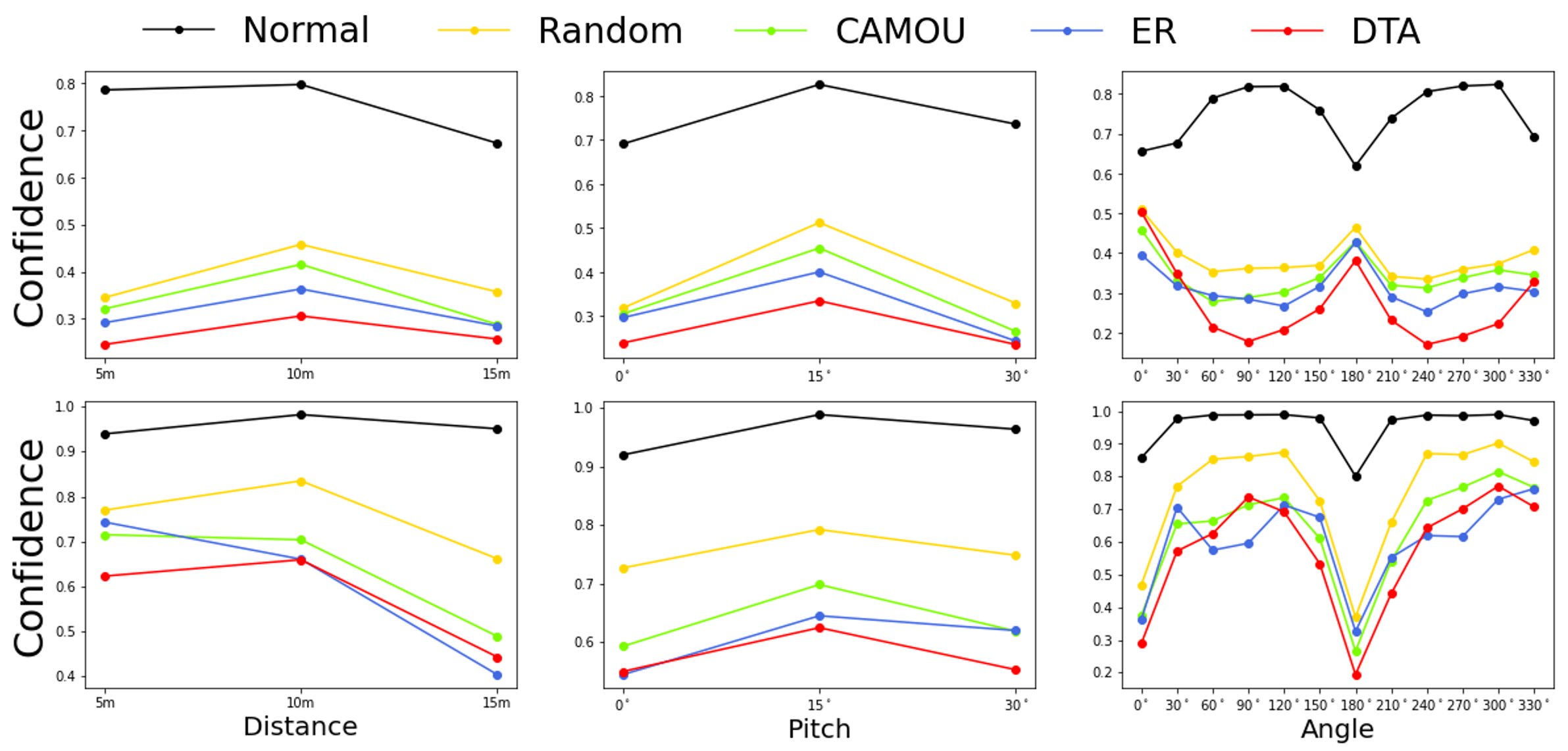}
        \caption{Target Average Confidence Score}
    \end{subfigure}
    \caption{Adversarial camouflage comparison on CARLA Simulator with different camera poses. First row: EfficientDetD0, second row YOLOv4}
    \label{fig:attack_pattern_eval}
\end{figure}

\begin{figure}[h]
    \centering
    \begin{subfigure}[b]{\textwidth}
        \centering
        \includegraphics[width=\textwidth]{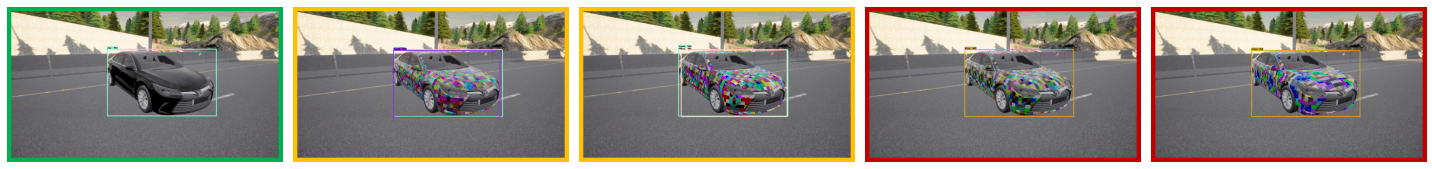}
    \end{subfigure}
    \begin{subfigure}[b]{\textwidth}
        \centering
        \includegraphics[width=\textwidth]{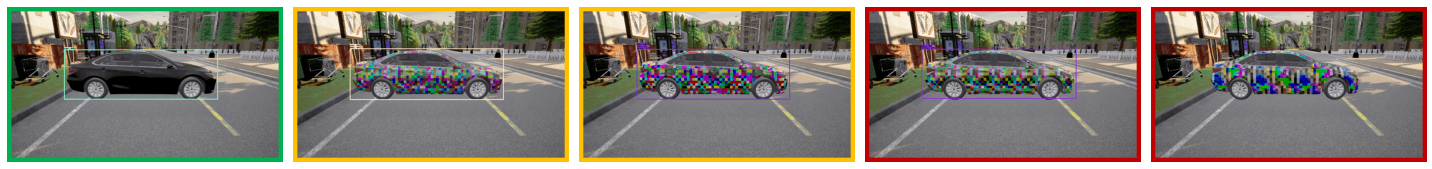}
    \end{subfigure}
     \begin{subfigure}[b]{\textwidth}
        \centering
        \includegraphics[width=\textwidth]{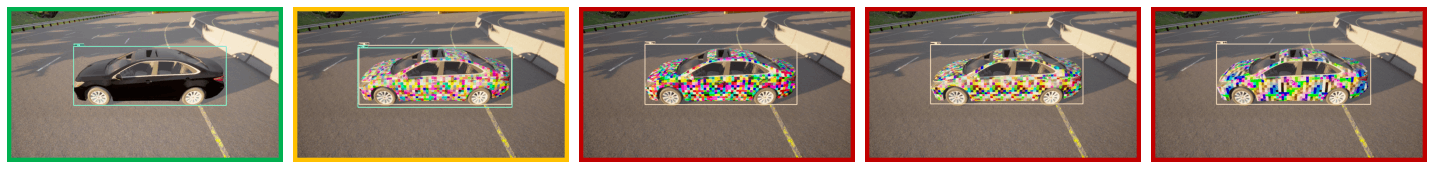}
    \end{subfigure}
     \begin{subfigure}[b]{\textwidth}
        \centering
        \includegraphics[width=\textwidth]{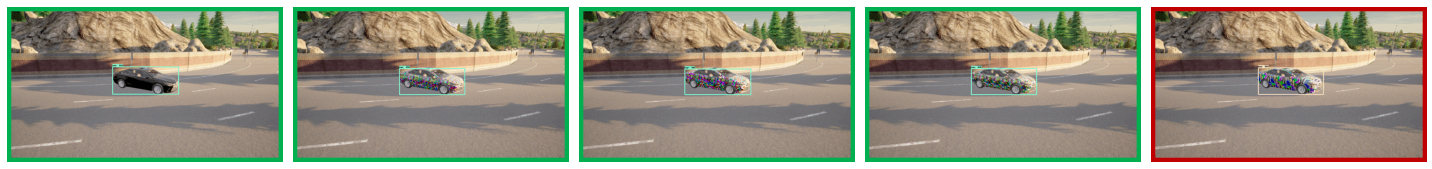}
    \end{subfigure}
    \begin{subfigure}[b]{\textwidth}
        \centering
        \includegraphics[width=\textwidth]{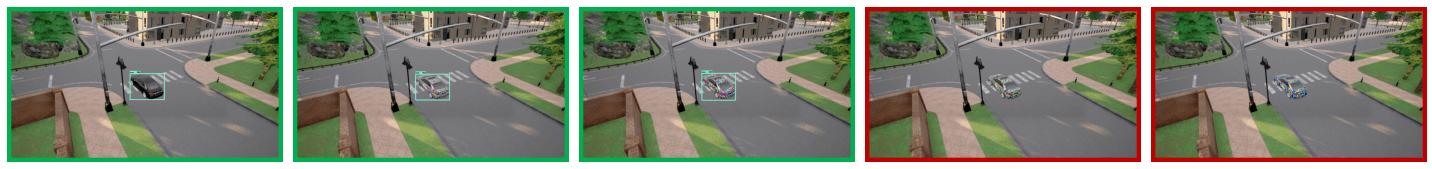}
    \end{subfigure}
    \begin{subfigure}[b]{\textwidth}
        \centering
        \includegraphics[width=\textwidth]{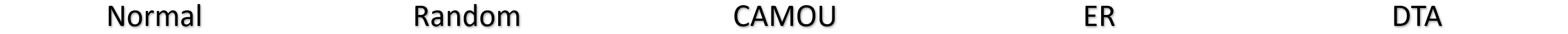}
    \end{subfigure}
    \caption{Sample images of adversarial camouflage on CARLA Simulator with different camera poses.
    Border: Red = miss-detection; Yellow = partially correct (other label also detected); Green = correct (detected as car).}
    \label{fig:target_model_sample}
\end{figure}

\textbf{Targeted Model Evaluation}
We perform comparative experiments to evaluate our adversarial camouflages with previous methods by attacking the COCO pre-trained EfficientdetD0 \cite{efficientdet} and YOLOv4 \cite{yolov4}.
Figure \ref{fig:attack_pattern_eval} describes adversarial camouflages comparison on different camera poses. From the figure, it can be inferred that our attack pattern is more robust with various transformations compared to other textures. Figure \ref{fig:target_model_sample} shows sample images of each adversarial camouflage on different camera poses. In the figures, red and green border represents miss-detection and correct detection, respectively, while yellow border signifies partially-correct detection (i.e., other labels are also detected).

\begin{figure}[h]
    \centering
    \includegraphics[width=0.95\textwidth]{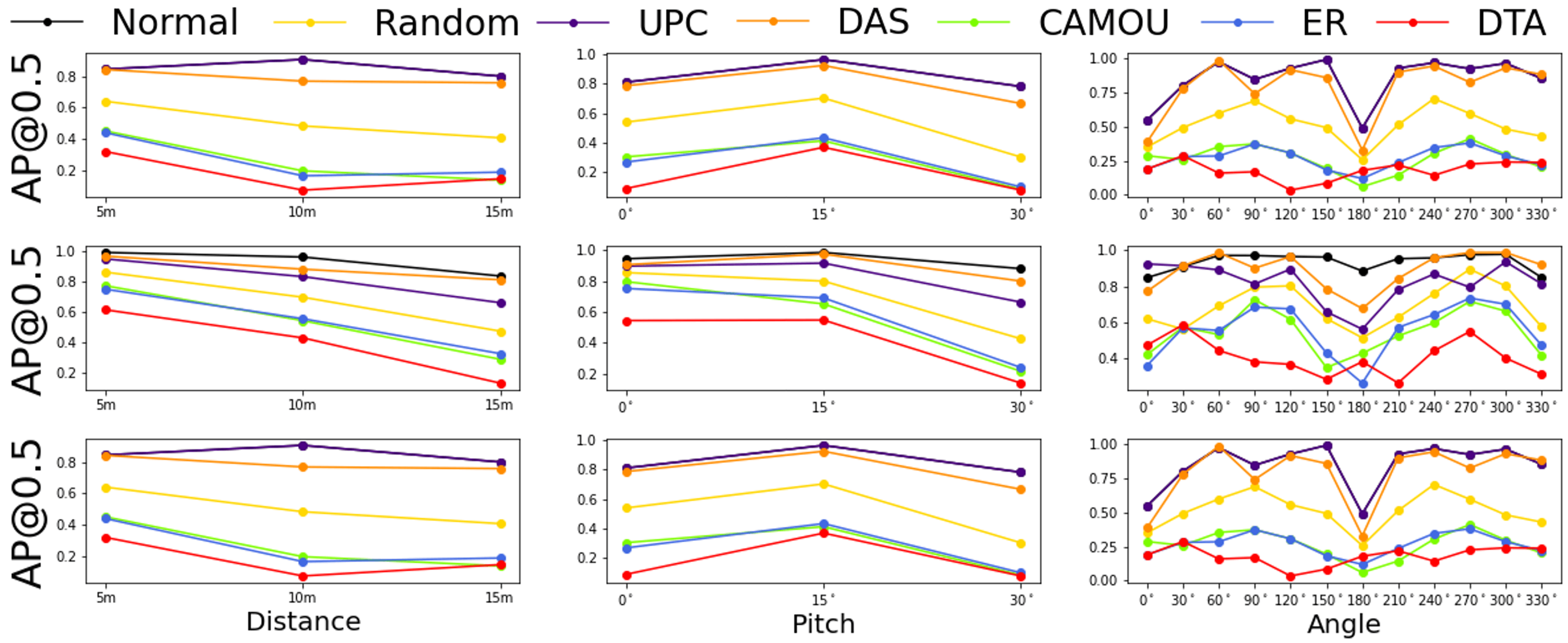}
    \caption{Attack transferability results by transformations. First row: SSD, second row: Faster R-CNN, third row: Mask R-CNN}
    \label{fig:transferability_by_transformation}
\end{figure}

\pagebreak
\textbf{Transferability Evaluation}
We further perform comparative experiments to evaluate our adversarial camouflages with previous methods in transferability setting which target object, transformations, and target model are different during the camouflage generation phase. The target models in this evaluation are COCO pre-trained SSD \cite{ssd}, Faster R-CNN \cite{fasterrcnn}, and Mask-RCNN \cite{maskrcnn}. 
Figure  \ref{fig:transferability_by_transformation} presents attack transferability evaluation results by transformation(distance, pitch and angle). From the figure, we can infer that our attack pattern is still more robust in the transferability setting compared to others.
Furthermore, Figure \ref{fig:transferability_sample} shows sample images of each adversarial camouflage on different camera poses and target model.

\begin{figure}[h]
    \centering
    \begin{subfigure}[b]{\textwidth}
        \centering
        \includegraphics[width=\textwidth]{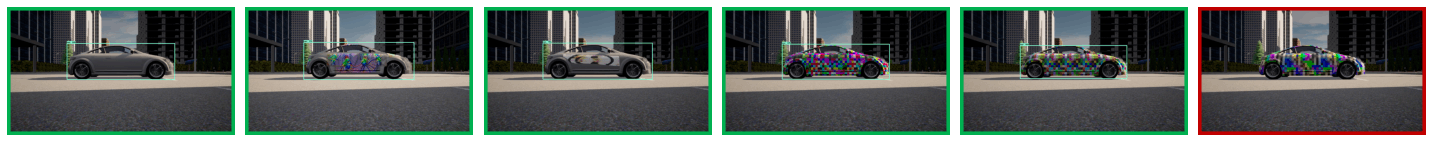}
    \end{subfigure}
    \begin{subfigure}[b]{\textwidth}
        \centering
        \includegraphics[width=\textwidth]{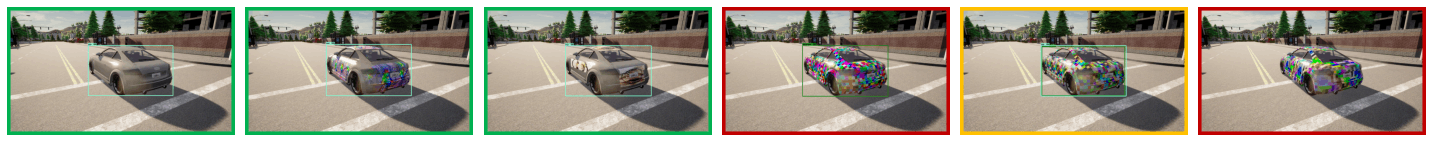}
    \end{subfigure}
    \begin{subfigure}[b]{\textwidth}
        \centering
        \includegraphics[width=\textwidth]{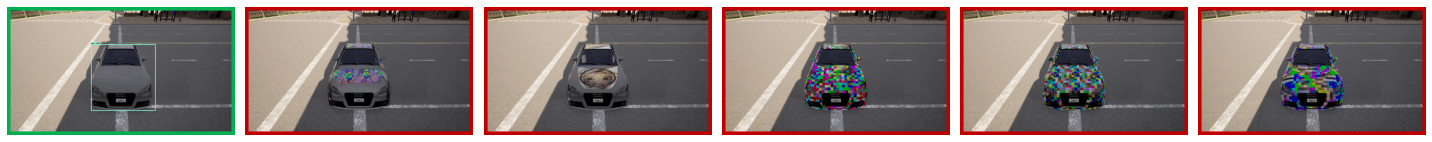}
    \end{subfigure}
    \begin{subfigure}[b]{\textwidth}
        \centering
        \includegraphics[width=\textwidth]{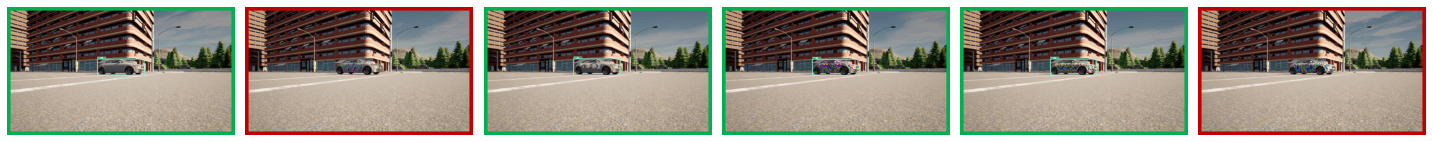}
    \end{subfigure}
     \begin{subfigure}[b]{\textwidth}
        \centering
        \includegraphics[width=\textwidth]{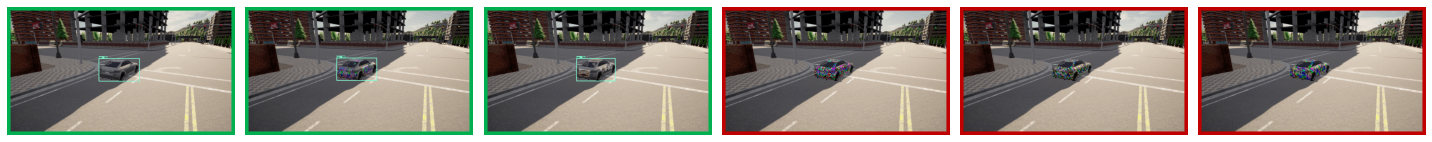}
    \end{subfigure}
    \begin{subfigure}[b]{\textwidth}
        \centering
        \includegraphics[width=\textwidth]{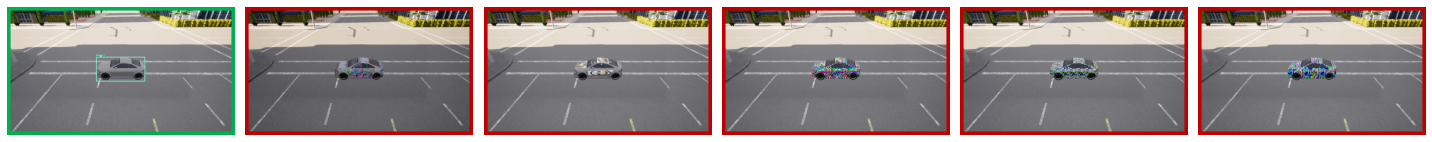}
    \end{subfigure}
     \begin{subfigure}[b]{\textwidth}
        \centering
        \includegraphics[width=\textwidth]{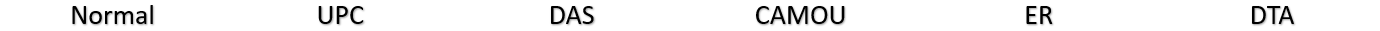}
    \end{subfigure}
\end{figure}

\begin{figure}[h] \ContinuedFloat
    \centering
    \begin{subfigure}[b]{\textwidth}
        \centering
        \includegraphics[width=\textwidth]{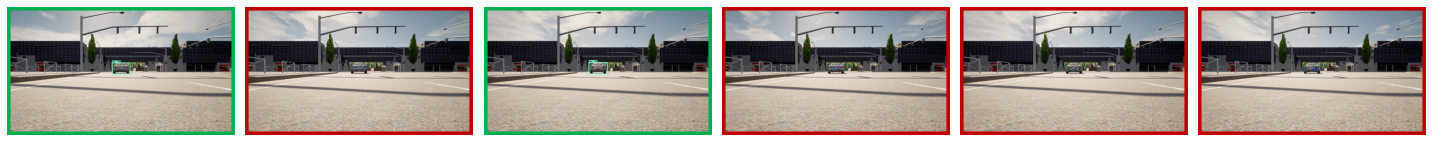}
    \end{subfigure}
    \begin{subfigure}[b]{\textwidth}
        \centering
        \includegraphics[width=\textwidth]{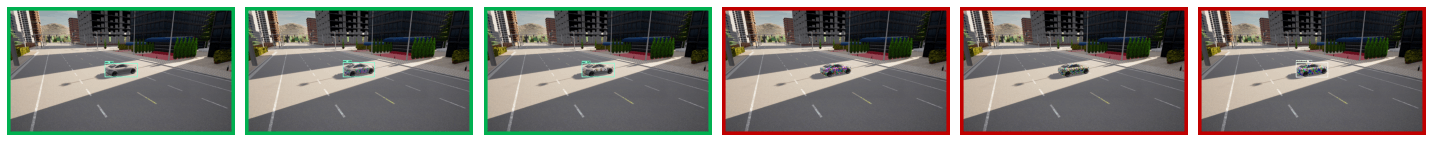}
    \end{subfigure}
    \begin{subfigure}[b]{\textwidth}
        \centering
        \includegraphics[width=\textwidth]{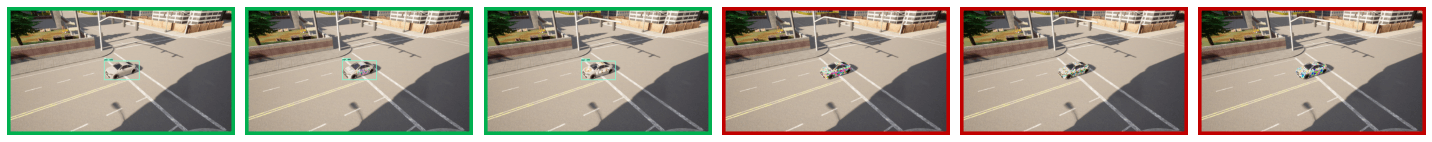}
    \end{subfigure}
     \begin{subfigure}[b]{\textwidth}
        \centering
        \includegraphics[width=\textwidth]{latex/final_sup_images/fig13_transferability_label.png}
        \caption{SSD model predictions}
    \end{subfigure}
\end{figure}

\begin{figure}[h] \ContinuedFloat
    \begin{subfigure}[b]{\textwidth}
        \centering
        \includegraphics[width=\textwidth]{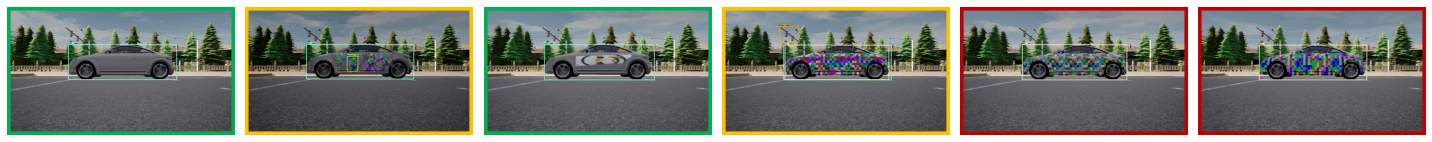}
    \end{subfigure}
    \begin{subfigure}[b]{\textwidth}
        \centering
        \includegraphics[width=\textwidth]{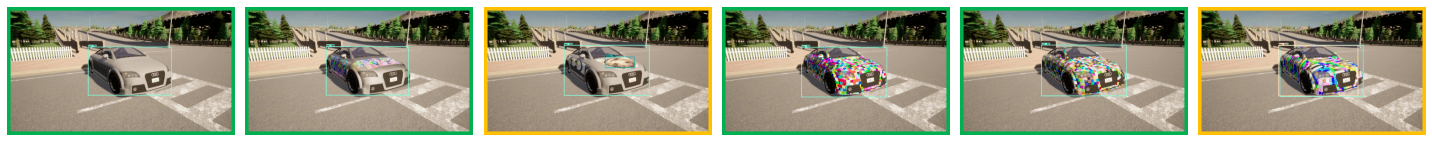}
    \end{subfigure}
    \begin{subfigure}[b]{\textwidth}
        \centering
        \includegraphics[width=\textwidth]{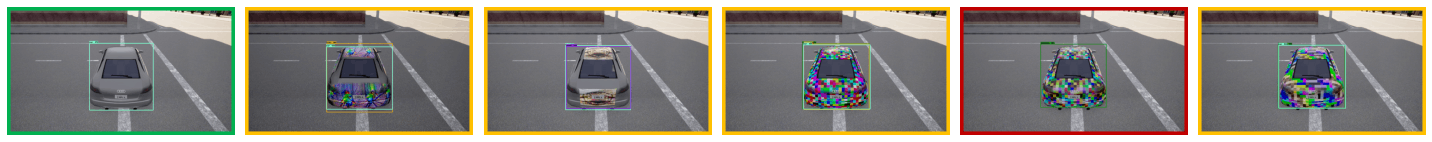}
    \end{subfigure}
    \begin{subfigure}[b]{\textwidth}
        \centering
        \includegraphics[width=\textwidth]{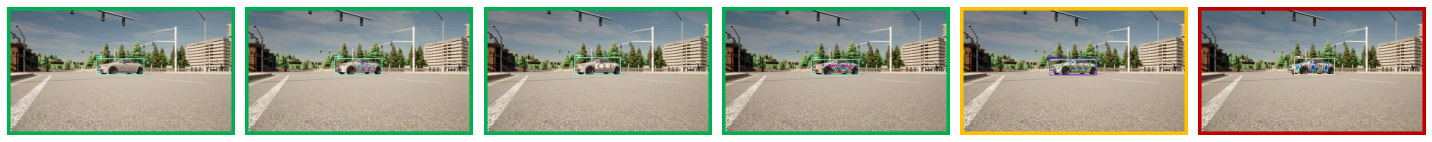}
    \end{subfigure}
    \begin{subfigure}[b]{\textwidth}
        \centering
        \includegraphics[width=\textwidth]{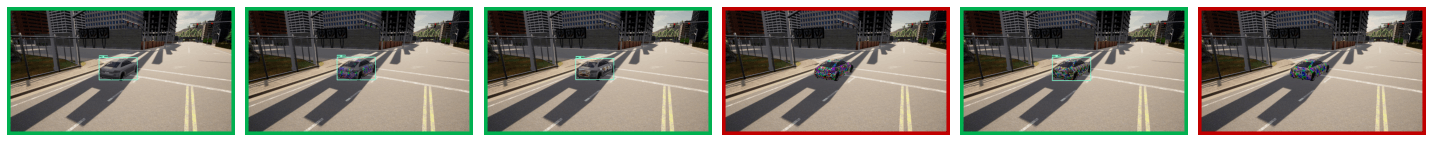}
    \end{subfigure}
     \begin{subfigure}[b]{\textwidth}
        \centering
        \includegraphics[width=\textwidth]{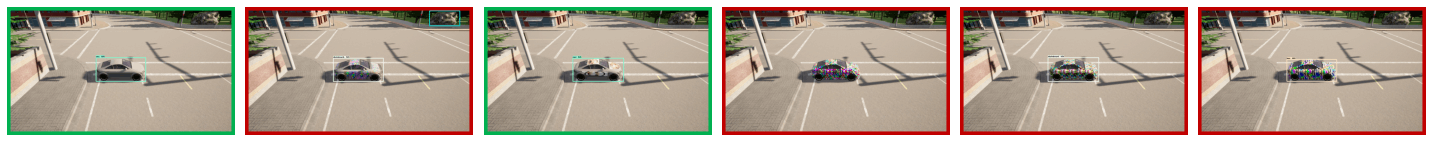}
    \end{subfigure}
    \begin{subfigure}[b]{\textwidth}
        \centering
        \includegraphics[width=\textwidth]{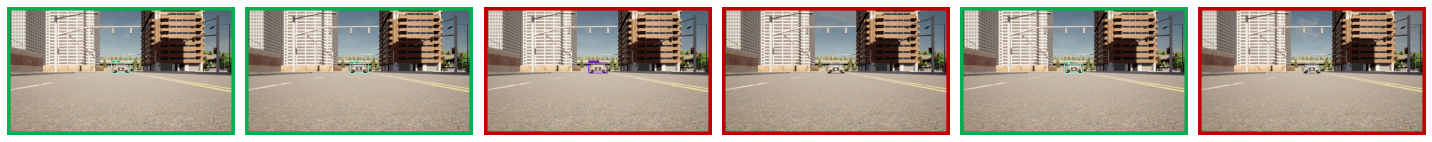}
    \end{subfigure}
     \begin{subfigure}[b]{\textwidth}
        \centering
        \includegraphics[width=\textwidth]{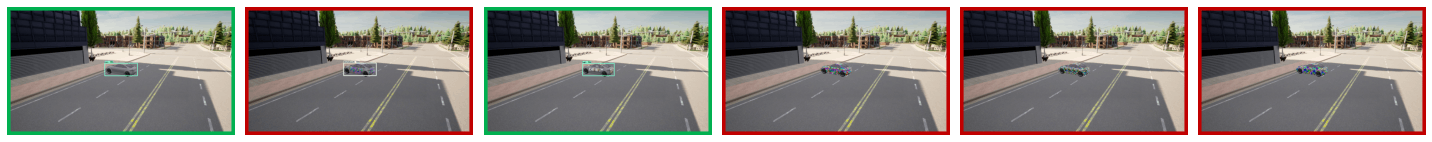}
    \end{subfigure}
      \begin{subfigure}[b]{\textwidth}
        \centering
        \includegraphics[width=\textwidth]{latex/final_sup_images/fig13_transferability_label.png}
        \caption{Faster R-CNN model predictions}
    \end{subfigure}
\end{figure}



\begin{figure}[h] \ContinuedFloat
    \begin{subfigure}[b]{\textwidth}
        \centering
        \includegraphics[width=\textwidth]{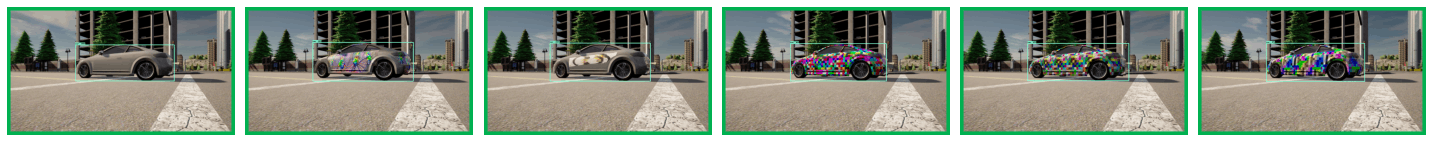}
    \end{subfigure}
    \begin{subfigure}[b]{\textwidth}
        \centering
        \includegraphics[width=\textwidth]{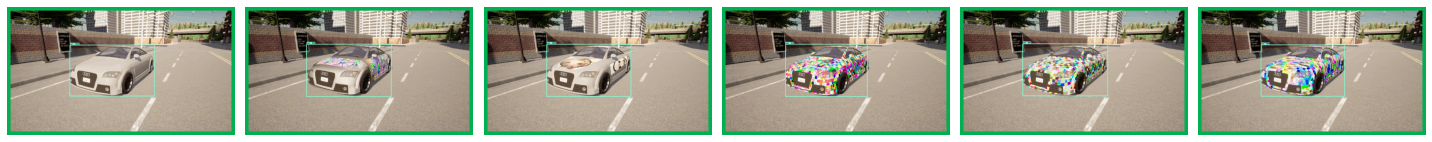}
    \end{subfigure}
    \begin{subfigure}[b]{\textwidth}
        \centering
        \includegraphics[width=\textwidth]{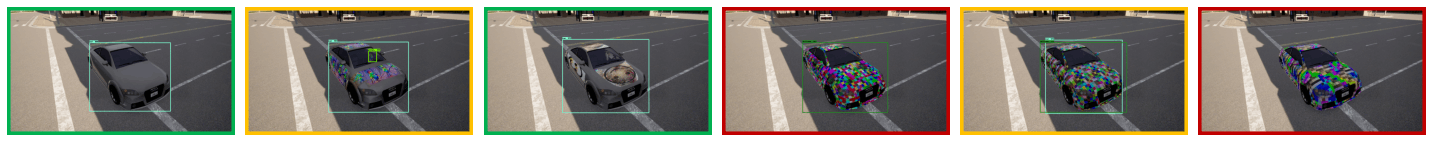}
    \end{subfigure}
    \begin{subfigure}[b]{\textwidth}
        \centering
        \includegraphics[width=\textwidth]{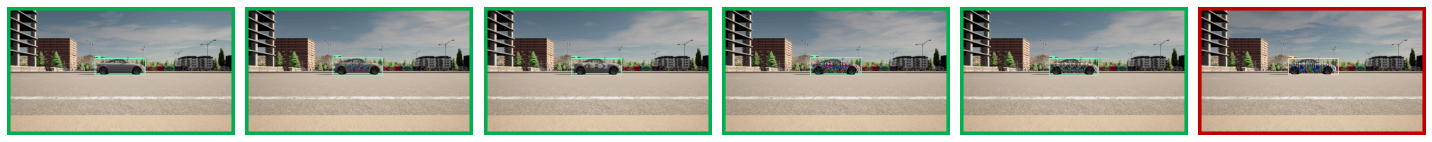}
    \end{subfigure}
    \begin{subfigure}[b]{\textwidth}
        \centering
        \includegraphics[width=\textwidth]{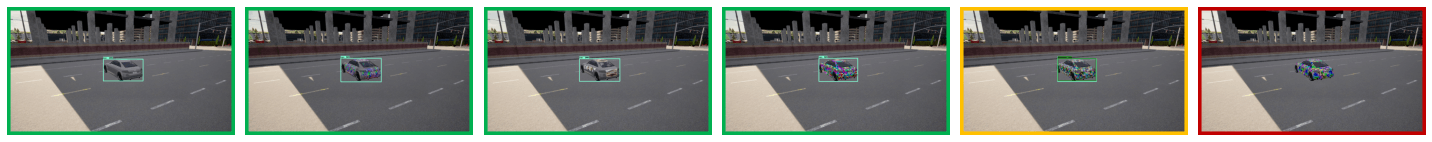}
    \end{subfigure}
    \begin{subfigure}[b]{\textwidth}
        \centering
        \includegraphics[width=\textwidth]{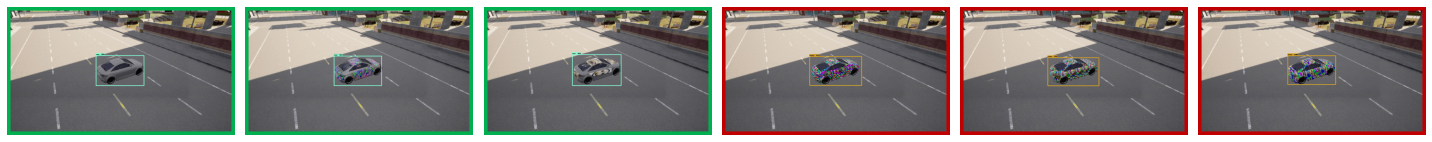}
    \end{subfigure}
    \begin{subfigure}[b]{\textwidth}
        \centering
        \includegraphics[width=\textwidth]{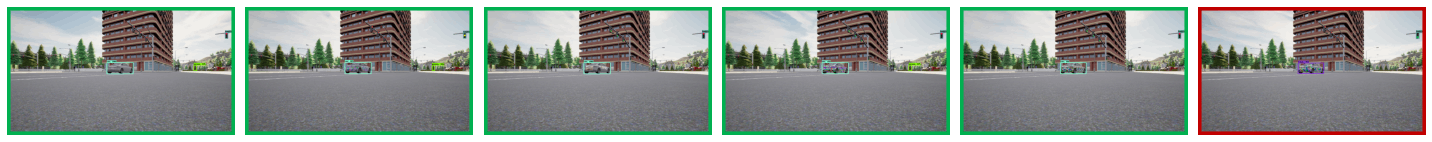}
    \end{subfigure}
     \begin{subfigure}[b]{\textwidth}
        \centering
        \includegraphics[width=\textwidth]{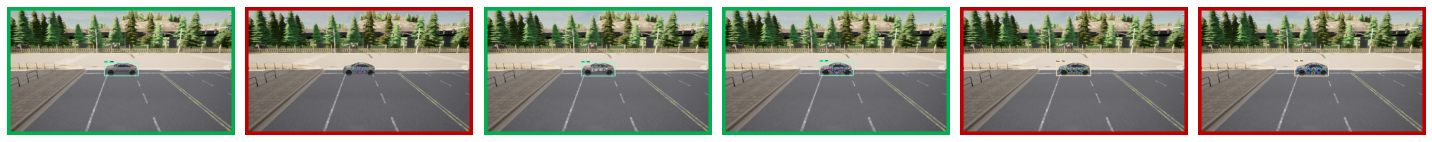}
    \end{subfigure}
     \begin{subfigure}[b]{\textwidth}
        \centering
        \includegraphics[width=\textwidth]{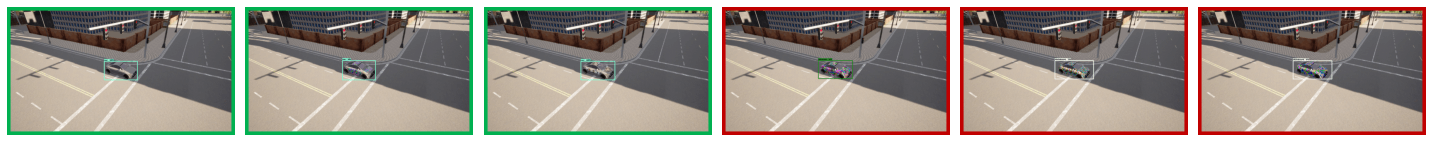}
    \end{subfigure}
     \begin{subfigure}[b]{\textwidth}
        \centering
        \includegraphics[width=\textwidth]{latex/final_sup_images/fig13_transferability_label.png}
        \caption{Mask R-CNN model predictions}
    \end{subfigure}
    \caption{Transferability: Sample images of adversarial camouflage on CARLA Simulator with different camera poses.
    Border: Red = miss-detection; Yellow = partially correct (other label also detected); Green = correct (detected as car).}
    \label{fig:transferability_sample}
\end{figure}

\pagebreak
\textbf{Detail Visualization of our Camouflage on Simulated Environment}
We provide more detailed samples for showing how our camouflage performance under a variety of camera poses and locations in a simulated environment.
Figure \ref{fig:attack_pattern_eval_on_simulator} shows the evaluation of our camouflage pattern on CARLA Simulator, by varying value of pitch, camera distance, and for every thirty degree rotations.

\begin{figure}[h]
    \centering
    \begin{subfigure}[b]{\textwidth}
        \centering
        \includegraphics[width=\textwidth]{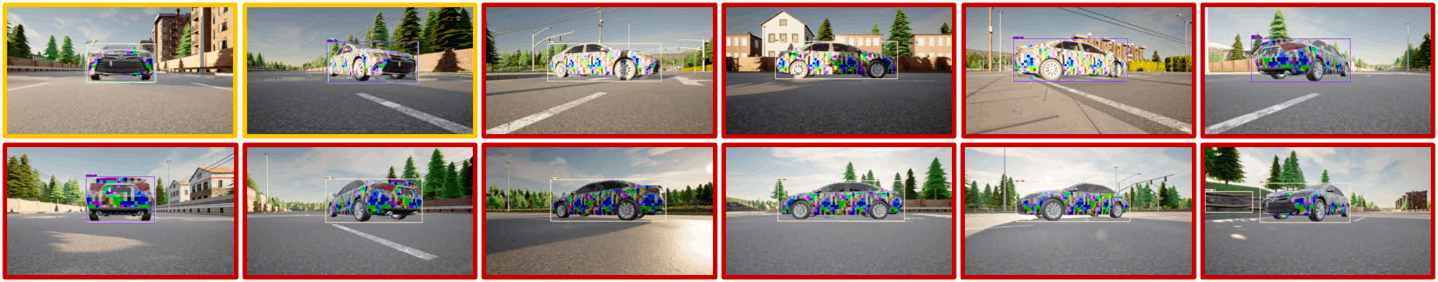}
        \caption{Pitch 0 deg, Distance 5m, Every 30 deg rotations}
        \label{fig:pitch_0_5m}
    \end{subfigure}
    \begin{subfigure}[b]{\textwidth}
        \centering
        \includegraphics[width=\textwidth]{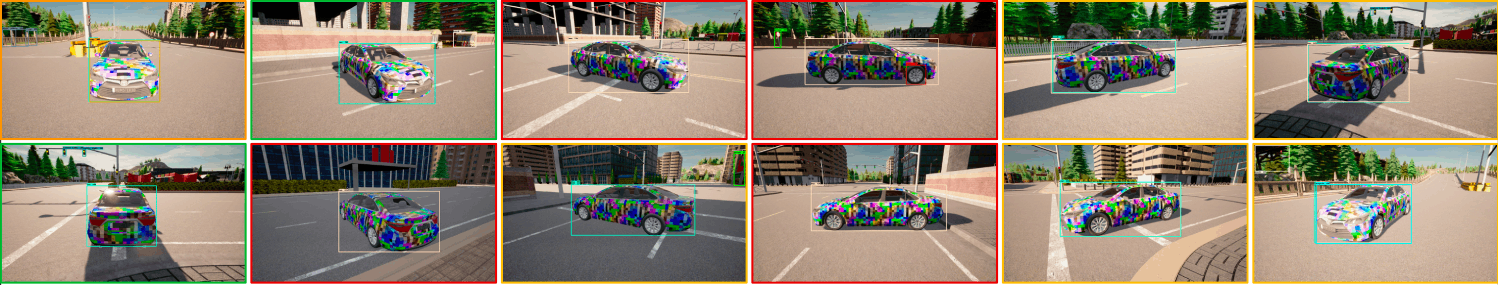}
        \caption{Pitch 15 deg, Distance 5m, Every 30 deg rotations}
        \label{fig:pitch_15_5m}
    \end{subfigure}
    \begin{subfigure}[b]{\textwidth}
        \centering
        \includegraphics[width=\textwidth]{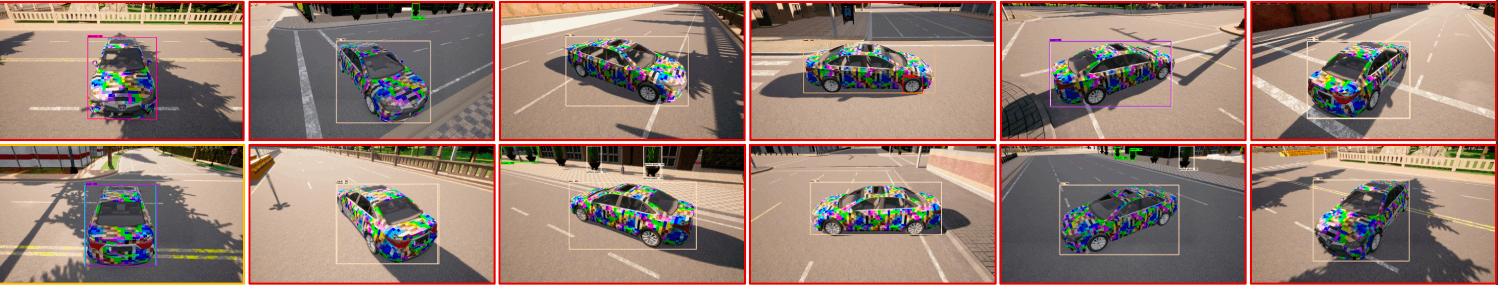}
        \caption{Pitch 30 deg, Distance 5m, Every 30 deg rotations}
        \label{fig:pitch_30_5m}
    \end{subfigure}
    \begin{subfigure}[b]{\textwidth}
        \centering
        \includegraphics[width=\textwidth]{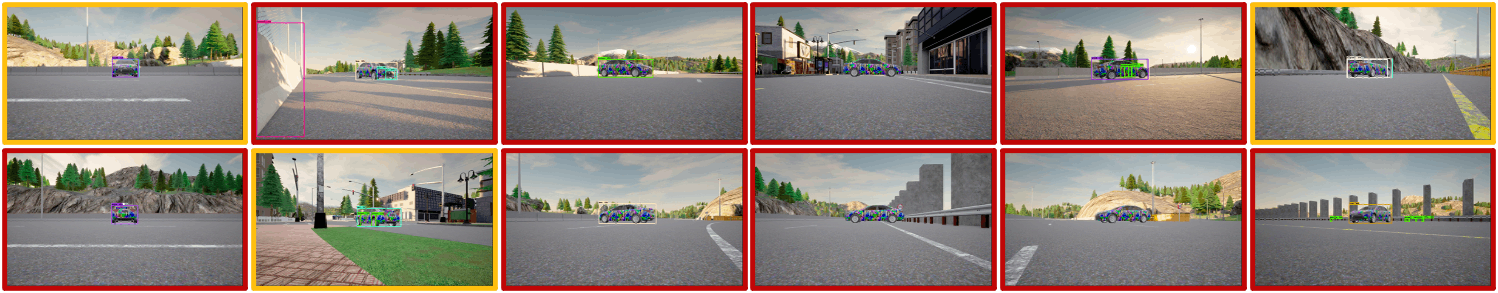}
        \caption{Pitch 0 deg, Distance 10m, Every 30 deg rotations}
        \label{fig:pitch_0_10m}
    \end{subfigure}
    \begin{subfigure}[b]{\textwidth}
        \centering
        \includegraphics[width=\textwidth]{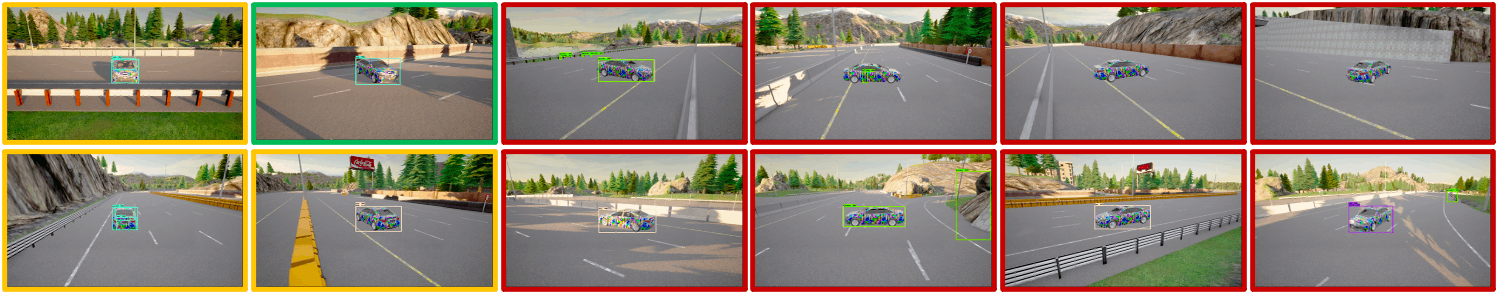}
        \caption{Pitch 15 deg, Distance 10m, Every 30 deg rotations}
        \label{fig:pitch_15_10m}
    \end{subfigure}
    \begin{subfigure}[b]{\textwidth}
        \centering
        \includegraphics[width=\textwidth]{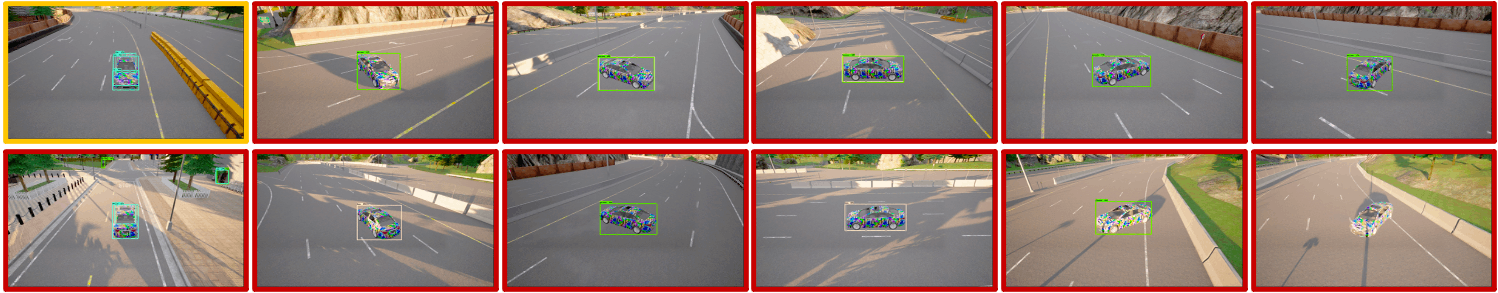}
        \caption{Pitch 30 deg, Distance 10m, Every 30 deg rotations}
        \label{fig:pitch_30_10m}
    \end{subfigure}
    \label{fig:attack_pattern_eval_on_simulator}
\end{figure}

\begin{figure}[ht] \ContinuedFloat
    \centering
    \begin{subfigure}[b]{\textwidth}
        \centering
        \includegraphics[width=\textwidth]{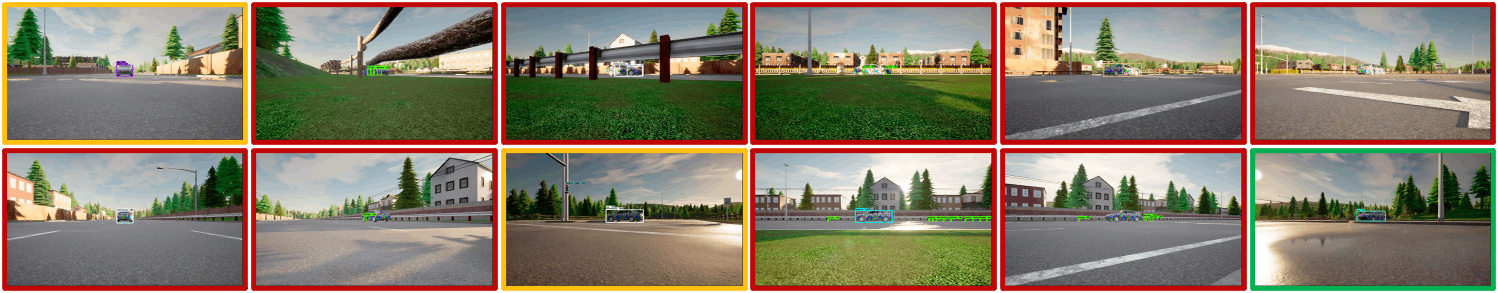}
        \caption{Pitch 0 deg, Distance 15m, Every 30 deg rotations}
        \label{fig:pitch_0_15m}
    \end{subfigure}
    \begin{subfigure}[b]{\textwidth}
        \centering
        \includegraphics[width=\textwidth]{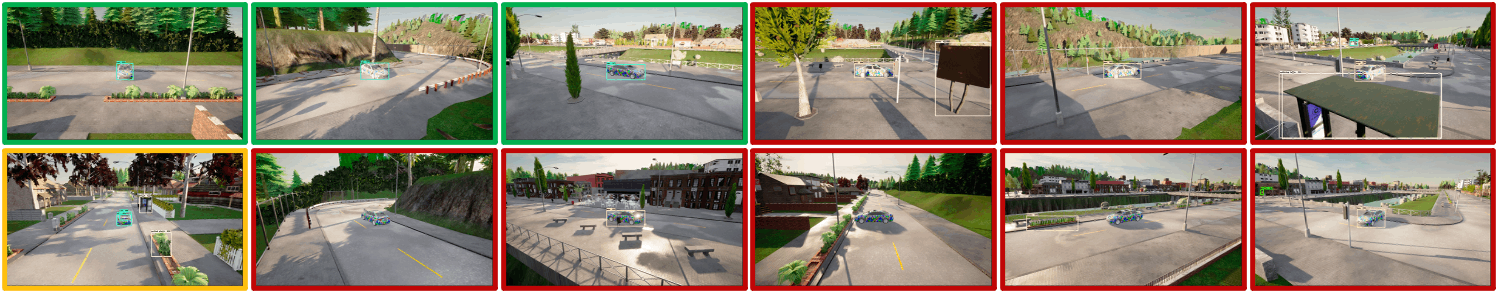}
        \caption{Pitch 15 deg, Distance 15m, Every 30 deg rotations}
        \label{fig:pitch_15_15m}
    \end{subfigure}
    \begin{subfigure}[b]{\textwidth}
        \centering
        \includegraphics[width=\textwidth]{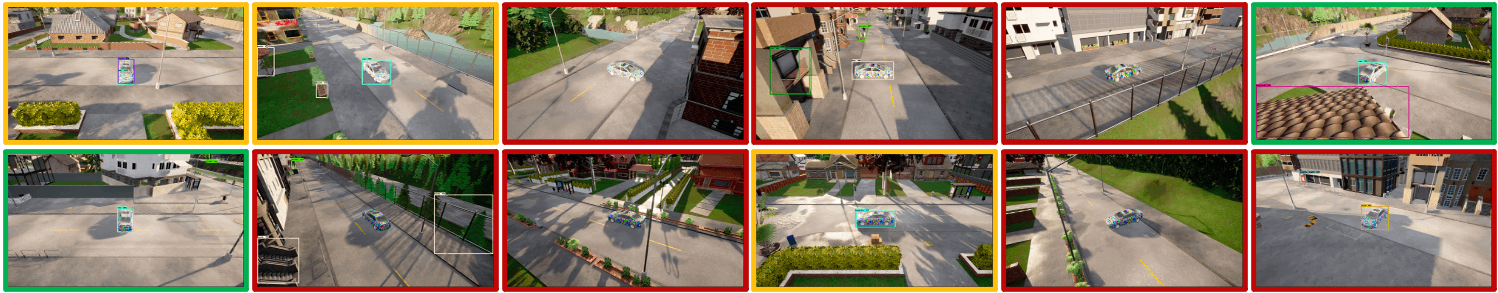}
        \caption{Pitch 30 deg, Distance 15m, Every 30 deg rotations}
        \label{fig:pitch_30_15m}
    \end{subfigure}
    \caption{Our Adversarial Camouflage Evaluation on CARLA Simulator. Border: Red = miss-detection; Yellow = partially correct (other label also detected); Green = correct.}
    \label{fig:attack_pattern_eval_on_simulator}
\end{figure}

\begin{figure}
    \centering
    \begin{subfigure}[b]{\textwidth}
        \centering
        \includegraphics[width=\textwidth]{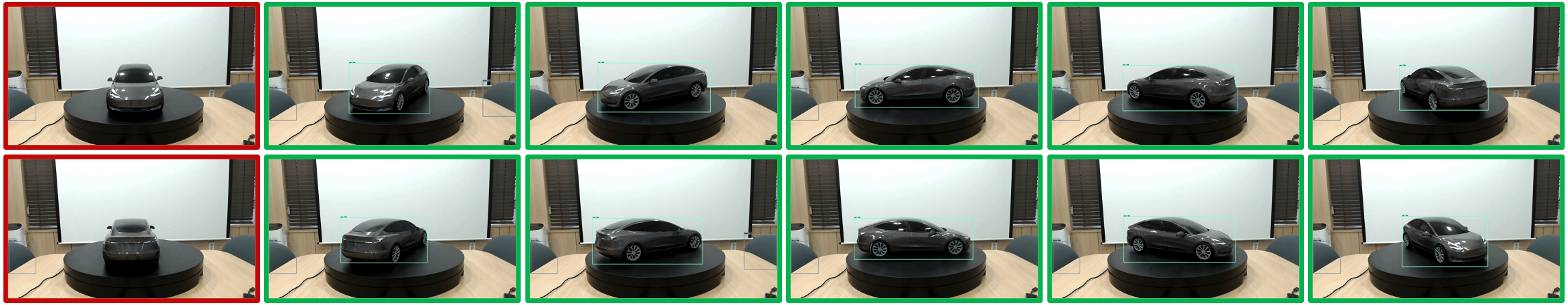}
        \caption{Normal Tesla Model 3 Evaluation}
        \label{fig:normal_tesla}
    \end{subfigure}
    \begin{subfigure}[b]{\textwidth}
        \centering
        \includegraphics[width=\textwidth]{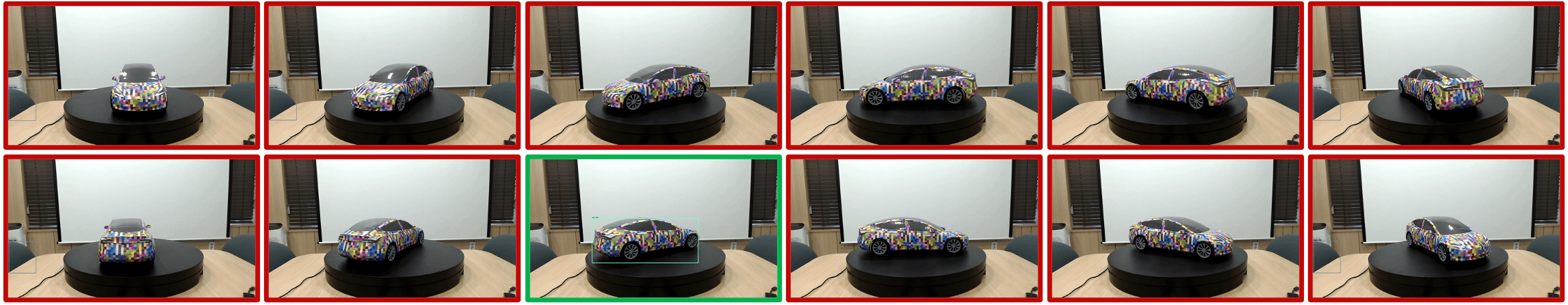}
        \caption{Attacked Tesla Model 3 Evaluation}
        \label{fig:attack_tesla}
    \end{subfigure}
    \caption{Our Adversarial Camouflage Evaluation on Real World. Border: Red = miss-detection; Yellow = partially correct (other label also detected); Green = correct.}
    \label{fig:attack_pattern_eval_on_real_world}
\end{figure}

\textbf{Detail Visualization in the Real World}
We provide more detailed samples for showing how our camouflage performance in the real world.
Figure \ref{fig:attack_pattern_eval_on_real_world} shows our evaluations on a lifelike miniature of Tesla Model 3. In particular, Figure \ref{fig:normal_tesla} shows the detection for normal situation (i.e., normal version of Tesla Model 3), whereas Figure \ref{fig:attack_tesla} shows the detection when adversarial attack is in operation.

\pagebreak

\clearpage

\end{document}